%% file: root.tex
\documentclass[journal]{IEEEtran}
\ifCLASSINFOpdf
  \usepackage[pdftex]{graphicx}
  \graphicspath{../figs}
  \DeclareGraphicsExtensions{.pdf,.jpeg,.png}
\else
\fi
\usepackage{hyperref}
\usepackage{booktabs}
\usepackage{hyperref}
\usepackage{dirtree}
\usepackage{caption}
\usepackage{multirow}
\usepackage{makecell}
\usepackage{pifont}
\usepackage[table,dvipsnames]{xcolor}
\usepackage{subcaption}
\usepackage{amsmath,amssymb,amsfonts}

\input{acronyms}
\newcommand{\cmark}{\textcolor{ForestGreen}{\ding{51}}} 
\newcommand{\xmark}{\textcolor{red}{\ding{55}}} 

\begin{document}
%
\title{The NavINST Dataset for Multi-Sensor Autonomous Navigation}
%
%
%
\author{{Paulo Ricardo Marques de Araujo}\textsuperscript{\textsection},
        {Eslam Mounier}\textsuperscript{\textsection},
        {Qamar Bader},
        {Emma Dawson},
        {Shaza I. Kaoud Abdelaziz},
        {Ahmed Zekry},
        {Mohamed Elhabiby} and
        {Aboelmagd Noureldin}
\thanks{{\textsection} Authors contributed equally.}
\thanks{Paulo Ricardo Marques de Araujo, Qamar Bader, Emma Dawson, Ahmed Zekry, and Shaza I. Kaoud Abdelaziz were with the  Department of Electrical and Computer Engineering, Queen's University, Kingston, ON, Canada (e-mail: \{paulo.araujo, qamar.bader, emma.dawson, ahmed.zekry, shaza.kaoud\}@queensu.ca).}%
\thanks{Eslam Mounier was with the  Department of Electrical and Computer Engineering, Queen's University, Kingston, ON, Canada, and the Ain Shams University, Cairo, Egypt (e-mail: eslam.abdelmoneem@queensu.ca).}%
\thanks{Mohamed Elhabiby was with Micro Engineering Tech. Inc., Calgary, AB, Canada (email: elhabiby@microengineering.ca).}%
\thanks{Aboelmagd Noureldin was with the Department of Electrical and Computer Engineering, Royal Military College of Canada, Kingston, ON, Canada (email: aboelmagd.noureldin@rmc.ca).}%
}

\maketitle

\begin{abstract}
The \ac{navinst} Laboratory has developed a comprehensive multisensory dataset from various road-test trajectories in urban environments, featuring diverse lighting conditions, including indoor garage scenarios with dense 3D maps. This dataset includes multiple commercial-grade IMUs and a high-end tactical-grade IMU. Additionally, it contains a wide array of perception-based sensors, such as a solid-state LiDAR—making it one of the first datasets to do so—a mechanical LiDAR, four electronically scanning RADARs, a monocular camera, and two stereo cameras. The dataset also includes forward speed measurements derived from the vehicle's odometer, along with accurately post-processed high-end GNSS/IMU data, providing precise ground truth positioning and navigation information. The NavINST dataset is designed to support advanced research in high-precision positioning, navigation, mapping, computer vision, and multisensory fusion. It offers rich, multi-sensor data ideal for developing and validating robust algorithms for autonomous vehicles. Finally, it is fully integrated with the \ac{ros}, ensuring ease of use and accessibility for the research community. The complete dataset and development tools are available at \textcolor{blue}{\textbf{\href{https://navinst.github.io}{navinst.github.io}}}.
\end{abstract}


%
\IEEEpeerreviewmaketitle

\setlength{\tabcolsep}{3pt}
\acresetall
\section{Introduction}
The last decade has witnessed significant advancements in autonomous driving, robotics, and computer vision, transforming these fields with innovative applications. In particular, \acp{av} technology has ushered in a new era of transportation, promising increased safety, efficiency, and convenience \cite{autonmousfuture}. These advancements in \acp{av} are fundamentally reliant on robust navigation systems capable of achieving higher levels of autonomy by operating seamlessly in diverse and dynamic environments while ensuring accuracy, reliability, and adaptability \cite{reid2019localization}.

A major enabler of these research advances has been the publication of diverse datasets by research groups that provide high-quality standardized data that supports the development, testing, and benchmarking of innovative algorithms. Datasets such as the widely renowned KITTI dataset \cite{geiger_vision_2013} have become foundational resources for tasks such as odometry, object detection, and sensor fusion, supporting thousands of research works around the world, overcoming limitations related to data accessibility and system constraints.
After KITTI, more datasets were published introducing diverse sensor configurations, research focuses and challenges. For instance, the Oxford RobotCar dataset \cite{maddern_1_2017} tackles large-scale challenges with over 1000 km of multi-sensory data, the Complex Urban dataset \cite{jeong_complex_2018} focuses \ac{lidar} technology in complex urban settings, the Mulran dataset \cite{kim_mulran_2020} focuses on structural place recognition, UrbanNav \cite{hsu_hong_2023} focuses on \ac{gnss}-challenged environments such as in deep urban areas, and Boreas \cite{burnett_boreas_2023} addresses multi-season challenges.

An integral component of these impactful datasets is the accompanying scientific paper that provides a detailed description of the dataset. The scope of dataset papers extends beyond simply presenting raw data. Their innovation lies in the thoughtful design, utility, and comprehensive documentation they provide, as well as the lessons learned throughout the production of the dataset. These documents enable researchers to seamlessly utilize the data for algorithm development and testing while allowing the research community to replicate, adapt, and extend the work. This, in turn, facilitates the creation of new systems and datasets to bridge research gaps, enhancing data quality and availability to fulfill the growing variety of research needs.

\begin{figure}
    \centering
    \includegraphics[width=\columnwidth]{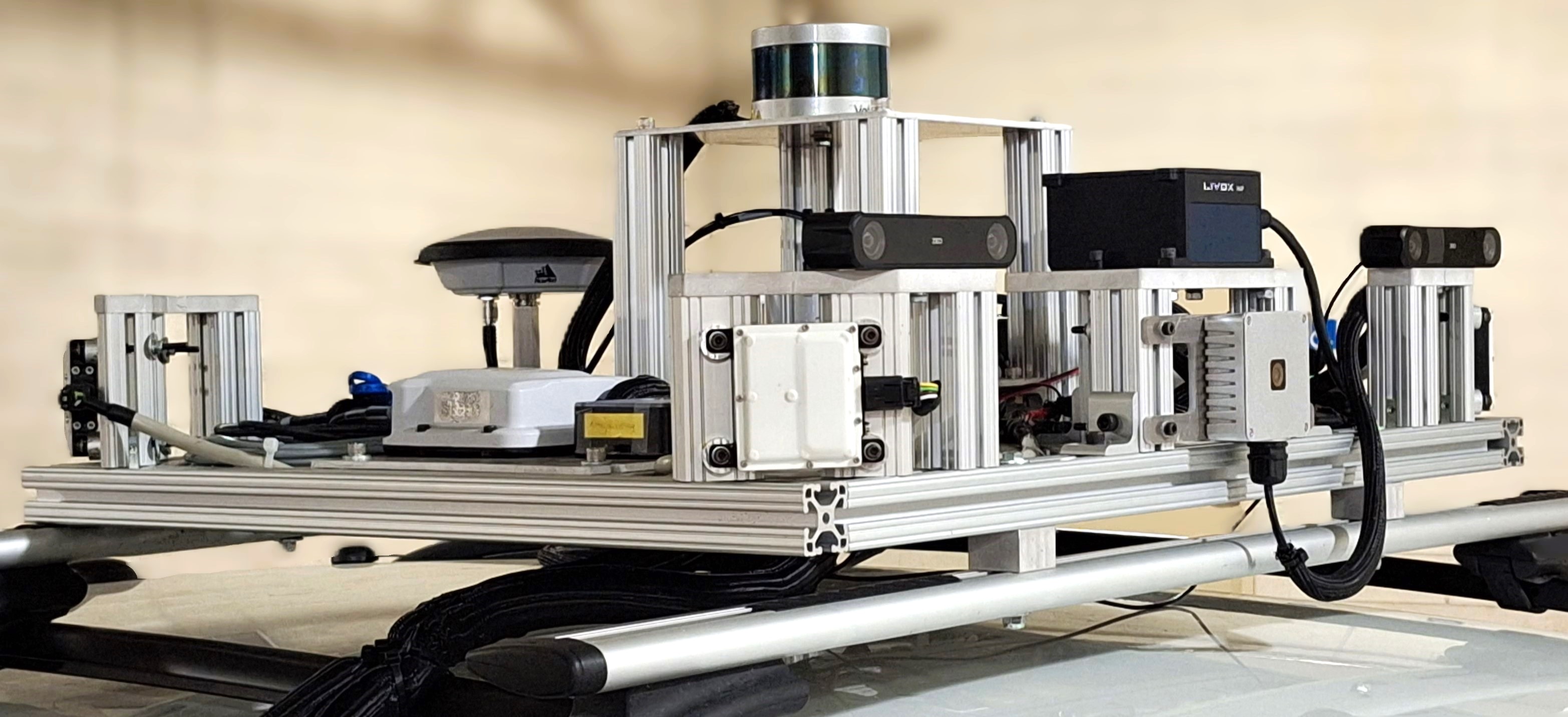}
    \caption{\ac{navinst} multisensory platform.}
    \label{fig:platform}
\end{figure}

In this line of work, we have developed the \ac{navinst} dataset, exemplifying thoughtful dataset design by providing a comprehensive collection of multi-modal sensor data tailored for research in high-precision positioning, navigation, mapping, and multi-sensor fusion. Constructed using \ac{ros}, the dataset was collected in Kingston, ON, and Calgary, AB, Canada, spanning approximately 80 kilometers of diverse urban and environmental conditions. It features repeated trajectories under varied lighting, traffic, and urban scenarios, as well as challenging indoor environments, making it a versatile resource for researchers, industry professionals, and developers in autonomous systems and navigation. The platform, illustrated in \figurename~\ref{fig:platform}, integrates a comprehensive sensor suite and a robust reference system, detailed in more detail in section~\ref{sec:platform}. This platform will be used to continue to perform road tests in various environments and in different cities. We will continue to update the published data with the newly collected data to support research activities in the area.  

The main contributions of the \ac{navinst} dataset are as follows:
\begin{itemize}
    \item We repeated trajectories under different lighting conditions and across diverse urban scenarios.
    \item We collected data in two indoor garages, and we also provided the 3D maps of these environments.
    \item Our dataset includes data from multiple commercial \acp{imu}, useful for advanced navigation-based research.
    \item Our dataset features a solid-state \ac{lidar}, making it one of the first datasets to provide such data.
    \item The dataset features multiple \acp{radar}, including four \acp{esr} providing 360$^\circ$ coverage around the vehicle, and a Doppler \ac{radar} mounted on the bumper for forward speed estimation. 
    \item We post-processed the \ac{gnss}/\ac{imu} data to provide accurate ground truth pose information for our outdoor trajectories.
    \item Our dataset is fully integrated with \ac{ros} ensuring convenient accessibility and usability, but a toolkit is also provided for those working offline.
\end{itemize}

The paper is structured as follows: Section \ref{sec:related-works} presents a literature survey of the existing positioning datasets. Section~\ref{sec:platform} presents the multisensory platform outlining the details of the sensor suite, including a description of the software architecture, reference system, and calibration. Section~\ref{sec:trajectories} explains important information about data collection, and section~\ref{sec:dataset-format} describes how the dataset is structured. Section~\ref{sec:demonstrations} presents some tests and demonstrations performed with the dataset, and section~\ref{sec:conclusion} concludes the paper.

\section{Related Works} \label{sec:related-works}
{
\setlength{\tabcolsep}{3pt}
\begin{table*}
\centering
\renewcommand{\arraystretch}{0.8}
\caption{Related Positioning Datasets. MC: Monocular Camera. SC: Stereo Camera. SS: Solid-state \ac{lidar}. A: Automotive 4D \ac{radar}. N: $360^\circ$ Navtech radar. T: Tactical-grade \ac{imu}. C: Commercial-grade \ac{imu}. GT: Ground truth pose source. RTK (Real-Time Kinematic) uses a \ac{gnss} base station and differential measurements to improve the estimations. RTX uses data from a global network of tracking stations to calculate corrections.}
\label{tab:datasets}
\resizebox{\textwidth}{!}{%
\begin{tabular}{@{}llllllccc@{}}
\toprule
\textbf{Name} &
  \textbf{Camera} &
  \textbf{\ac{lidar}} &
  \textbf{\ac{radar}} &
  \textbf{\ac{imu}} &
  \textbf{GT} &
  \textbf{Indoor} &
  \textbf{Night} &
 \textbf{\ac{ros} Support} \\ \midrule
KITTI (odometry)~\cite{geiger_vision_2013} &
  4$\times$ MC &
  1$\times$ 64 ch &
  \xmark &
  1$\times$ T &
  \ac{gnss}/\ac{imu} + RTK &
  \xmark &
  \xmark &
  \xmark \\
Oxford RobotCar~\cite{maddern_1_2017} &
  1$\times$ SC + 3$\times$ MC &
  1$\times$ 4 ch + 2$\times$ 1 ch &
  \xmark &
  1$\times$ T &
  \ac{gnss}/\ac{imu} &
  \xmark &
  \cmark &
  \xmark \\
Oxford Radar RobotCar~\cite{barnes_oxford_2020} &
  1$\times$ SC + 3$\times$ MC &
  2$\times$ 32 ch + 2$\times$ 1 ch &
  1$\times$ N &
  1$\times$ T &
  \ac{gnss}/\ac{imu} + VO &
  \xmark &
  \xmark &
  \xmark \\
UrbanNav~\cite{hsu_hong_2023} &
  1$\times$ SC &
  1$\times$ 32 ch + 2$\times$ 16 ch &
  \xmark &
  1$\times$ C &
  \ac{gnss}/\ac{imu} + RTK &
  \textcolor{blue}{\textbf{!}} &
  \cmark &
  \cmark \\
Complex Urban~\cite{jeong_complex_2018} &
  \xmark &
  2$\times$ 16 ch + 2$\times$ 1 ch &
  \xmark &
  1$\times$ T + 1$\times$ C &
  \ac{gnss}/\ac{imu} &
  \textcolor{blue}{\textbf{!}} &
  \xmark &
  \cmark \\
MSC-RAD4R~\cite{choi_msc-rad4r_2023} &
  1$\times$ SC &
  1$\times$ 128 ch &
  1$\times$ A &
  1$\times$ C &
  \ac{gnss} + RTK &
  \textcolor{blue}{\textbf{!}} &
  \cmark &
  \cmark \\
MulRan~\cite{kim_mulran_2020} &
  \xmark &
  1$\times$ 64 ch &
  1$\times$ N &
  \xmark &
  SLAM &
  \xmark &
  \xmark &
  \cmark \\
K-Radar~\cite{paek_k-radar_2023} &
  4$\times$ SC &
  1$\times$ 128 ch + 1$\times$ 64 ch &
  1$\times$ A &
  2$\times$ C &
  \ac{gnss} + RTK* &
  \xmark &
  \cmark &
  \xmark \\
Boreas~\cite{burnett_boreas_2023} &
  1$\times$ MC &
  1$\times$ 128 ch &
  1$\times$ N &
  1$\times$ T &
  \ac{gnss}/\ac{imu} + RTX &
  \xmark &
  \cmark &
  \xmark \\
PixSet~\cite{deziel_pixset_2021} &
  4$\times$ MC &
  1$\times$ SS + 1$\times$ 64 ch &
  1$\times$ A &
  1$\times$ T &
  \ac{gnss}/\ac{imu} + RTK &
  \xmark &
  \cmark &
  \xmark \\
\textbf{\ac{navinst}} &
  2$\times$ SC + 1$\times$ MC &
  1$\times$ SS + 1$\times$ 16 ch &
  4$\times$ A + 1$\times$ 1D &
  4$\times$ C + 1$\times$ T &
  \acs{gnss}/\ac{imu} + RTK &
  \cmark &
  \cmark &
  \cmark \\ \bottomrule
  \multicolumn{9}{l}{\scriptsize \textcolor{blue}{\textbf{!}}~tunnels in some of the trajectories. * no GT reference is mentioned in the paper, this may be the most accurate information available.}
\end{tabular}%
}
\end{table*}
}

Several datasets have been published in recent years to enable research in positioning and navigation through the fusion of perception sensors, onboard motion sensors, and \ac{gnss}. \tablename~\ref{tab:datasets} summarizes a collection of these datasets.

One of the earliest contributions to this body of work is the KITTI odometry dataset \cite{geiger_vision_2013}. It consists of data from four monocular cameras, a 64-channel \ac{lidar}, as well as inertial and \ac{gnss} data. Additionally, all data is collected during daylight hours. More recent contributions to open navigation datasets include the Oxford RobotCar \cite{maddern_1_2017} and UrbanNav \cite{hsu_hong_2023} datasets. Both provide camera, \ac{lidar} and inertial measurement data. The Oxford RobotCar explores nighttime (darkened) driving, and UrbanNav provides \ac{ros} support to users. Neither dataset includes \ac{radar} data. The Complex Urban \cite{jeong_complex_2018} dataset provides neither \ac{radar} nor camera data but is interesting in that it provides both tactical and commercial grade \ac{imu} data and offers \ac{ros} support to facilitate its use. 

\ac{radar} technology has improved in the past decade, making it a promising sensor for positioning and navigation applications. As such, only a subset of existing datasets offer \ac{radar} data within their sensor suite. The datasets including \ac{radar} data are divided by the \ac{radar} technology used: mechanically scanning \ac{radar} or electronically scanning \ac{radar}. The Oxford Radar~\cite{barnes_oxford_2020}, MulRan~\cite{kim_mulran_2020}, and Boreas~\cite{burnett_boreas_2023} datasets provide data from a mechanically scanning $360^\circ$ \ac{radar}. The data provided by these \acp{radar} is high resolution, but the sensors are costly and do not provide Doppler velocity information \cite{kung-2021}. Other datasets employ \ac{esr} instead. A 3D \ac{esr} provides range, azimuth, and Doppler velocity measurements for target objects. A 4D \ac{esr} provides an additional elevation angle measurement. MSC-RAD4R~\cite{choi_msc-rad4r_2023} and K-Radar~\cite{paek_k-radar_2023} are interesting datasets, each providing data from a single 4D \ac{radar}. However, 4D \acp{radar} have limited fields of view, and a single \ac{radar} cannot provide a $360^\circ$ view of the vehicle's surroundings. The \ac{navinst} multisensory platform has four 4D \ac{esr}, enabling a $360^\circ$ view of the vehicle's environment and providing Doppler velocity information for all targets.  

The PixSet dataset published by Leddar Tech is among the first to provide data from a proprietary solid-state \ac{lidar} alongside a 64-channel scanning \ac{lidar} \cite{deziel_pixset_2021}. However, a single 3D \ac{radar} mounted at the front bumper of the data collection vehicle provides limited opportunities for \ac{radar}-based positioning research. Inertial data is restricted to a single tactical grade \ac{imu}. The \ac{navinst} dataset also provides a solid-state \ac{lidar} and a 16-channel mechanical \ac{lidar}, with the advantage of four 4D \ac{esr} for comparison and fusion opportunities.

\section{NavINST Multisensory Platform} \label{sec:platform}

In this section, we present the multisensory platform and its sensors, the \ac{ros} network, and the time synchronization strategy used to synchronize the multiple computers available in the vehicle. Also, we detail the post-processing approach to achieve accurate ground truth information and the intrinsic and extrinsic calibration of the perception sensors.

\subsection{Sensors} \label{sec:sensors}

\begin{figure}
    \centering
    \includegraphics[width=\columnwidth]{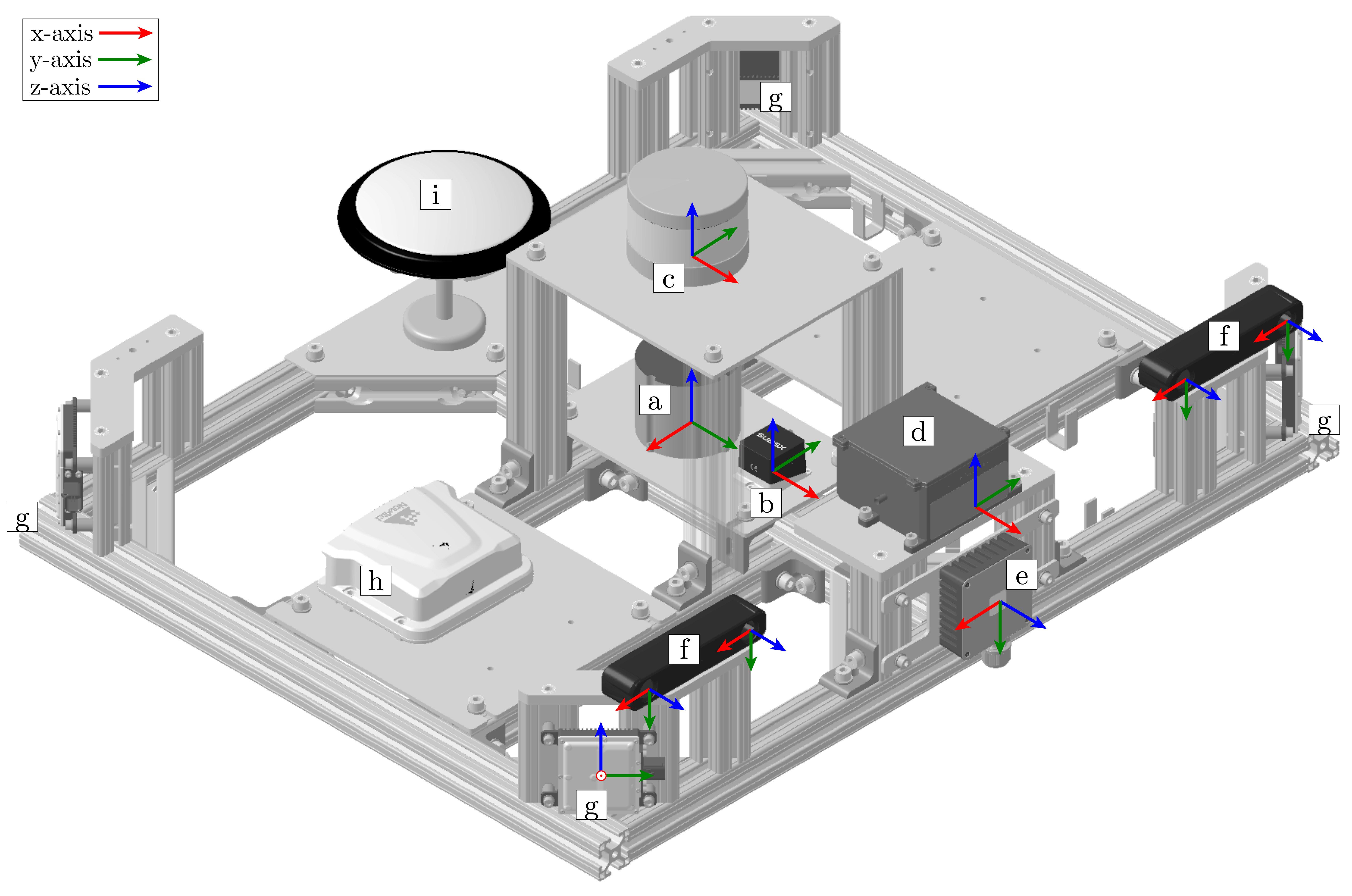}
    \caption{Sensors and their coordinate systems.}
    \label{fig:coordsys}
\end{figure}

The physical platform is shown in \figurename~\ref{fig:platform}, and the coordinate systems of the sensors are presented in \figurename~\ref{fig:coordsys}. The system is installed as a single unit on the vehicle's roof. This arrangement facilitates logistics and reduces the uncertainty of the extrinsic calibration. The data is stored in \ac{ros} \textit{bag} files, in which each sensor data is stored under a specific topic name. Sometimes, a sensor provides more than one piece of information, which demands multiple message types. Such message types can be found in the dataset repository as detailed in section~\ref{sec:dataset-format}. Specifications, topic names, and message types for sensors are summarized in \tablename~\ref{tab:setup-sensors}. 

\begin{table*}
\centering
\renewcommand{\arraystretch}{0.7}
\caption{Specifications, \ac{ros} topic names, and message types for sensors.}
\label{tab:setup-sensors}
\resizebox{\linewidth}{!}{%
\begin{tabular}{@{}c@{}c@{\hspace{5pt}}l@{}c@{\hspace{5pt}}cl@{\hspace{5pt}}l@{}}
\toprule
\textbf{Label} &
  \textbf{Sensor} &
  \textbf{Specifications} &
  \textbf{Qty} &
  \textbf{Hz} &
  \textbf{Topic Name} &
  \textbf{Message Type} \\ \midrule
a &
  \makecell[c]{\textbf{KVH1750}\\ {\acs{imu}}}  &
  \begin{tabular}[c]{@{}l@{}}\textbf{Gyroscope}\\ - Bias stability: $0.05~{}^\circ/\text{h}$\\ - Noise density: $0.012~{}^\circ/\sqrt{\text{h}}$\\ \textbf{Accelerometer}\\ - Bias stability: 7.5~mg\\ - Noise density: $0.0701~\text{m/s/}\sqrt{\text{h}}$\end{tabular} &
  1 &
  200 &
  /novatel/imu &
  sensor\_msgs/Imu \\ \midrule
b & 
\makecell[c]{\textbf{Xsens MTi-670G}\\ {\acs{imu}/\ac{gnss}}}  &
\begin{tabular}[c]{@{}l@{}}\textbf{Gyroscope}\\ - In-run bias stability: $8~{}^\circ/\text{h}$\\ - Noise density: $0.007~{}^\circ/s/\sqrt{\text{Hz}}$\\ \textbf{Accelerometer}\\ - In-run bias stability: 10 (x,y) 15 (z)~$\mu$g\\ - Noise density: $60~\mu\text{g/}\sqrt{\text{Hz}}$\\ \textbf{\acs{gnss} receiver}\\ - u-blox ZED F9\end{tabular}   &
  1 &
  100 &
  \makecell[l]{/xsens/imu\\/xsens/gnss\\/xsens/mag\\/xsens/pressure\\/xsens/temperature\\/xsens/filter/quaternion\\/xsens/filter/positionlla\\/xsens/filter/twist} &
  \makecell[l]{sensor\_msgs/Imu\\sensor\_msgs/NavSatFix\\geometry\_msgs/Vector3Stamped \\sensor\_msgs/FluidPressure\\sensor\_msgs/Temperature\\geometry\_msgs/QuaternionStamped\\geometry\_msgs/Vector3Stamped\\geometry\_msgs/TwistStamped} \\ \midrule
c &
  \makecell[c]{\textbf{Velodyne VLP-16} \\ 3D Mechanical\\\ac{lidar}} &
  \begin{tabular}[c]{@{}l@{}}
  HFOV: $360^\circ$\\ VFOV: $30^\circ$\\
  Angular resolution\\ - H: $0.1^\circ - 0.4^\circ$\\ - V: $2.0^\circ$\\
  \end{tabular} &
  1 &
  10 &
  /velodyne/lidar/points &
  sensor\_msgs/PointCloud2 \\ \midrule
d &
  \makecell[c]{\textbf{Livox HAP} \\3D Solid-state\\\ac{lidar}} &
  \begin{tabular}[c]{@{}l@{}}
  HFOV: $120^\circ$\\ VFOV: $25^\circ$\\ 
  Angular resolution\\ - H: $0.18^\circ$\\ - V: $0.23^\circ$\\
  \end{tabular} &  
  1 &
  10 &
  \makecell[l]{/livox/lidar/points\\/livox/lidar/imu} &
  \makecell[l]{sensor\_msgs/PointCloud2\\sensor\_msgs/Imu} \\ \midrule
e &
  \makecell[c]{\textbf{Luxonis OAK-1}\\\textbf{W PoE} \\ Monocular Camera} &
  \begin{tabular}[c]{@{}l@{}}DFOV: $150^\circ$\\ HFOV: $127^\circ$\\ VFOV: $79.5^\circ$\\ Resolution: 1MP ($1280 \times 720$) \\ Shutter: Global\end{tabular} &
  1 &
  30 &
  \makecell[l]{/oak/camera\_info\\/oak/image\_raw\_color/compressed} &
  \makecell[l]{sensor\_msgs/CameraInfo\\sensor\_msgs/CompressedImage} \\ \midrule
f &
\makecell[c]{\textbf{Stereolabs ZED X} \\ Stereo Camera} &
  \begin{tabular}[c]{@{}l@{}}DFOV: $120^\circ$\\ HFOV: $110^\circ$\\ VFOV: $80^\circ$\\ Resolution: 1MP ($960 \times 540$)\\ Shutter: Global\\ Baseline: 120 mm\end{tabular} &
  2 &
  15 &
  \makecell[l]{/zedx\_left/imu\\/zedx\_left/temperature\\/zedx\_left/camera\_left/camera\_info\\/zedx\_left/camera\_left/image\_raw\_color/compressed\\/zedx\_left/camera\_right/camera\_info\\/zedx\_left/camera\_right/image\_raw\_color/compressed\\
  /zedx\_right/imu\\/zedx\_right/temperature\\/zedx\_right/camera\_left/camera\_info\\/zedx\_right/camera\_left/image\_raw\_color/compressed\\/zedx\_right/camera\_right/camera\_info\\/zedx\_right/camera\_right/image\_raw\_color/compressed} &
  \makecell[l]{sensor\_msgs/Imu\\sensor\_msgs/Temperature\\sensor\_msgs/CameraInfo\\sensor\_msgs/CompressedImage\\sensor\_msgs/CameraInfo\\sensor\_msgs/CompressedImage\\
  sensor\_msgs/Imu\\sensor\_msgs/Temperature\\sensor\_msgs/CameraInfo\\sensor\_msgs/CompressedImage\\sensor\_msgs/CameraInfo\\sensor\_msgs/CompressedImage} \\ \midrule
g &
  \makecell[c]{\textbf{Smartmicro UMRR-96}\\\textbf{Type 153} \\4D \ac{esr}} &
  \begin{tabular}[c]{@{}l@{}}Range: 0.15 m ... 19.3 m\\ Range accuracy: \textless 0.15 m\\ Speed: -400 ... 140 km/h\\ Speed accuracy: \textless 0.15 m/s\\ HFOV: $130^\circ$\\ VFOV: $15^\circ$\\ Horizontal accuracy: $\leq 1^\circ$\\ Vertical accuracy: $\leq 2^\circ$\end{tabular} &
  4 &
  20 &
  \makecell[l]{/smartmicro/radar/front\_left\\/smartmicro/radar/front\_right\\/smartmicro/radar/rear\_left\\/smartmicro/radar/rear\_right} &
  sensor\_msgs/PointCloud2 \\ \midrule
h &
\makecell[c]{\textbf{Novatel PwrPak7-E1} \\ \acs{gnss}/\ac{imu} }&
  \begin{tabular}[c]{@{}l@{}}Constellations: GPS, GLONASS,\\ BeiDou, Galileo, IRNSS, SBAS,\\ QZSS, NavIC \\ Single point accuracy: 1.5 m\\ SBAS accuracy: 0.60 m\\ RTK accuracy: 0.01 m + 1 ppm \\ \textbf{\ac{imu}}: EPSON G320N \\ - Gyro Input Range: $\pm150{}^\circ/\text{s}$ \\ - Bias Repeatability: $0.5{}^\circ/\text{s}$ \\ - Angular Random Walk: $0.1{}^\circ/\sqrt{\text{h}}$ \\ - Accelerometer Range: $\pm 5$ g \\ - Accelerometer Bias Repeatability: 15 mg\end{tabular} &
  1 &
  \makecell[c]{50\\125} &
  \makecell[l]{/novatel/gps\\/novatel/Imu*} &
  \makecell[l]{gps\_common/GPSFix \\sensor\_msgs/Imu} \\ \midrule
i &
  \makecell[c]{\textbf{Novatel VEXXIS}\\\acs{gnss}-850} &
  \acs{gnss} Antenna &
  1 &
  - &
  - &
  - \\ \midrule
- &
  \makecell[c]{\textbf{OBDII}\\Automotive Scanner} &
  \begin{tabular}[c]{@{}l@{}}FORS-CAN ELM327 \\ - On-Board Diagnostics V2 Protocols \\ - USB2.0\end{tabular} &
  1 &
  16 &
  \makecell[l]{/obd2/speed\\/obd2/rpm\\/obd2/maf} &
  $\square$ \\ \midrule
- &
  \makecell[c]{\textbf{OPS241-A} \\ 1D Doppler \ac{radar}} &
  \begin{tabular}[c]{@{}l@{}}Range: 1 m ... 25 m\\ Maximum speed: 223 km/h\\ Speed resolution: 0.436 km/h\end{tabular} &
  1 &
  20 &
  /omnipresense/radar/front\_bumper &
  $\square$ \\ \bottomrule
  \multicolumn{7}{l}{$\square$ custom message. * available in indoor trajectories only.}
\end{tabular}%
}
\end{table*}

\begin{figure*}
    \centering
    \begin{subfigure}{0.325\textwidth}
        \centering
        \includegraphics[width=\textwidth]{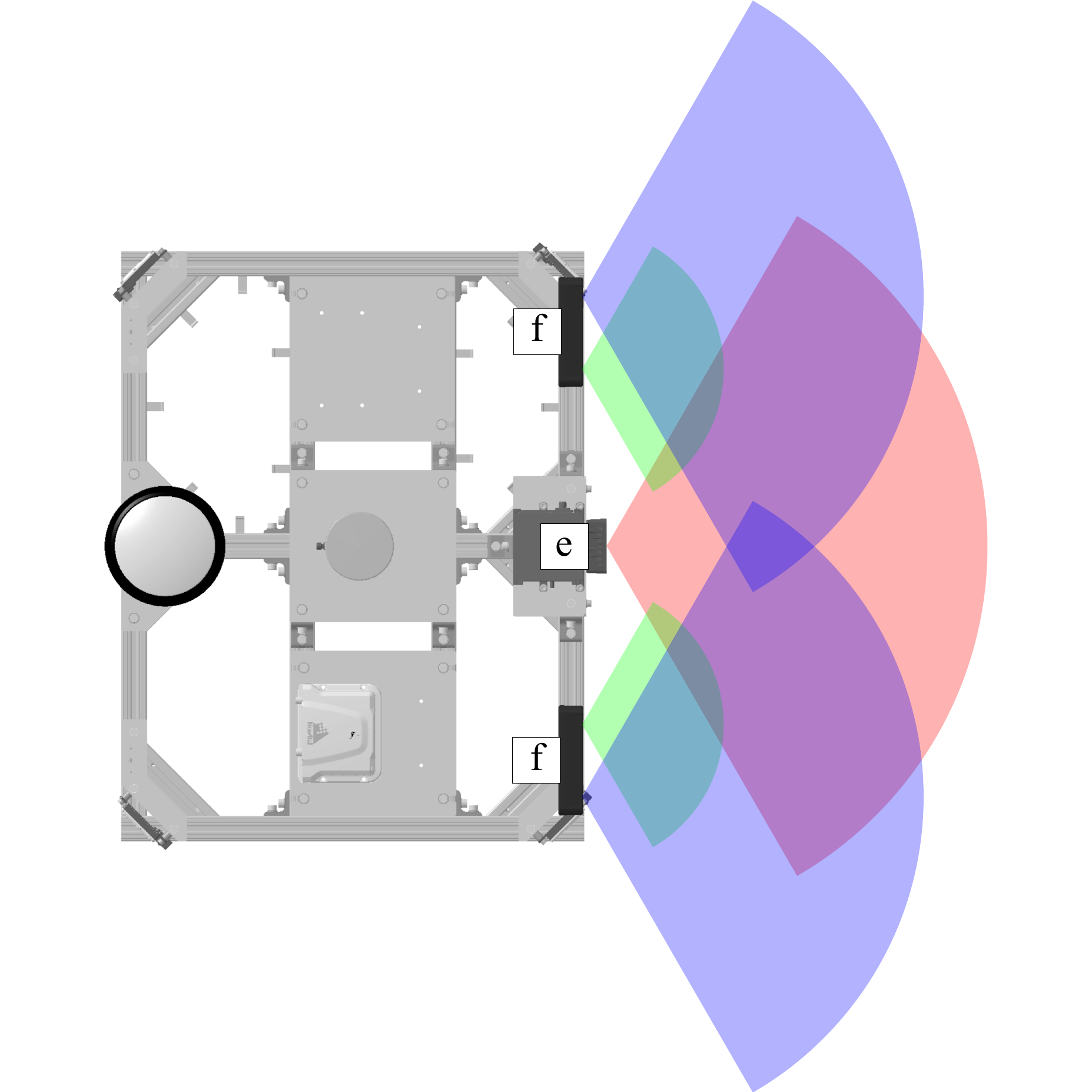}
        \caption{Cameras.}
        \label{fig:cameras}
    \end{subfigure}
    \begin{subfigure}{0.325\textwidth}
        \centering
        \includegraphics[width=\textwidth]{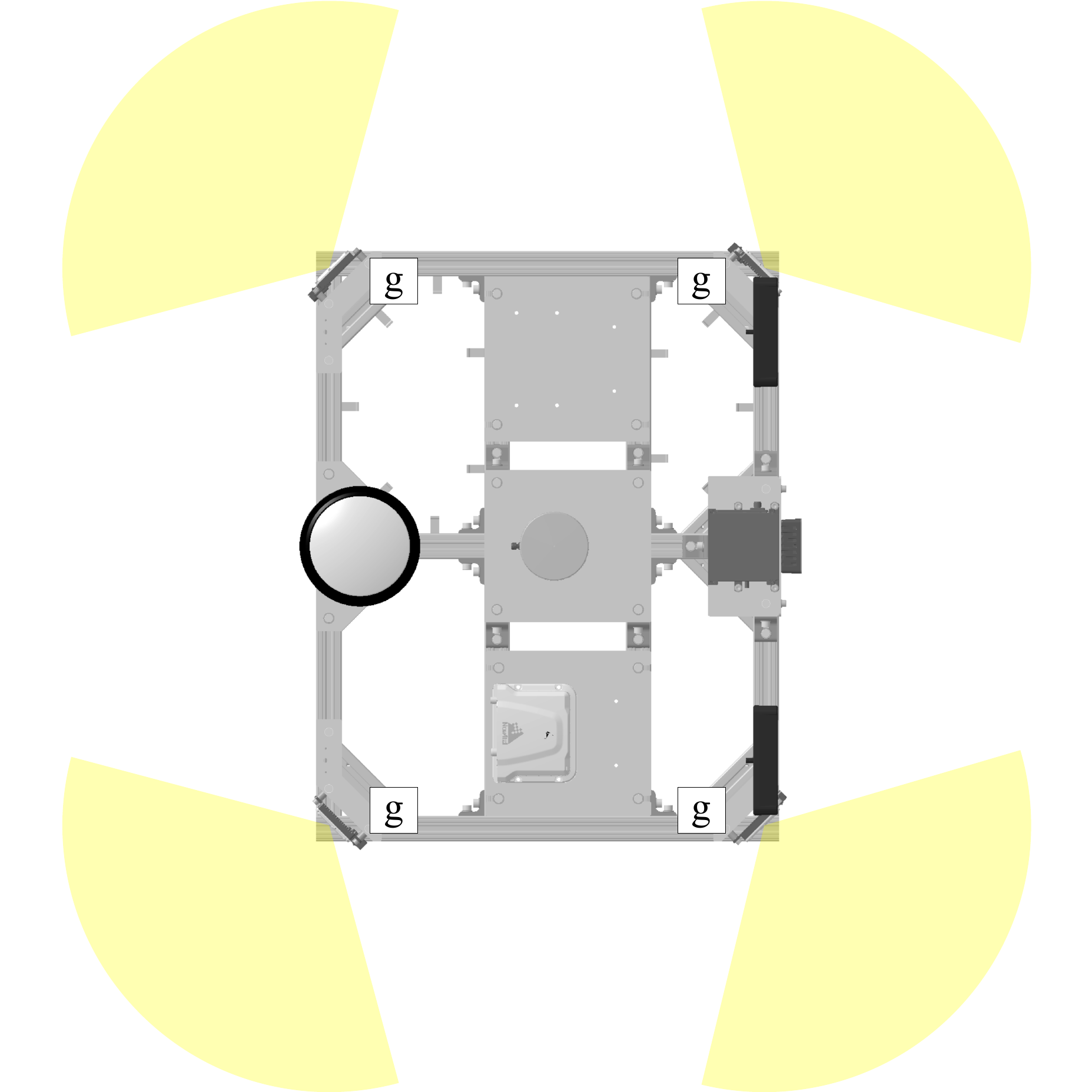}
        \caption{\acp{radar}.}
        \label{fig:radars}
    \end{subfigure}
    \begin{subfigure}{0.325\textwidth}
        \centering
        \includegraphics[width=\textwidth]{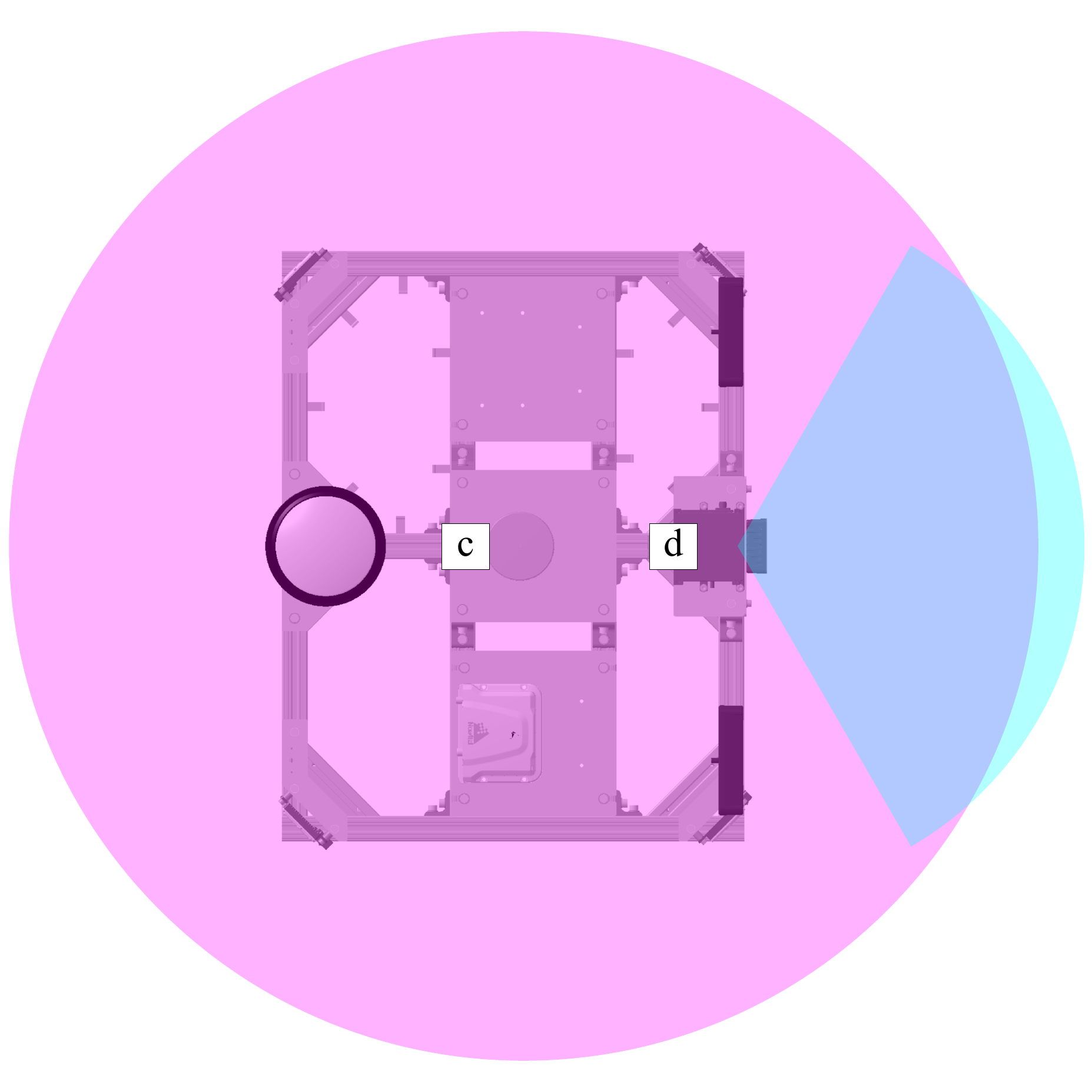}
        \caption{\acp{lidar}.}
        \label{fig:lidars}
    \end{subfigure}

    \caption{Schematic representation of the field of view of perception sensors.}
    \label{fig:fovs}
\end{figure*}

The platform was designed for 360$^\circ$ coverage, combining multiple sensors to achieve this objective, as shown in \figurename~\ref{fig:fovs}. The dataset includes cameras that are mounted in a configuration that can facilitate various investigations. For example, the multiple aligned cameras can be used to build different baselines for stereo vision applications.

Each stereo camera contains an \ac{imu}, but the manufacturer has not disclosed the model. Internal tests indicate that this \ac{imu} is of a commercial grade.
Additionally, the solid-state \ac{lidar} is equipped with a Bosch BMI088 \ac{imu}. Combined with the \ac{imu} inside the MTi-670G, the dataset includes four commercial \acp{imu}, providing ample opportunities for various investigations. In addition, our dataset includes two GNSS receivers, the Novatel PwrPak7-E1 and the Xsens MTi-670G, which can be used to investigate different fusion algorithms with both high-end and commercial-grade receivers.

\begin{figure*}
    \centering
    \begin{subfigure}{0.325\textwidth}
        \centering
        \includegraphics[width=\textwidth]{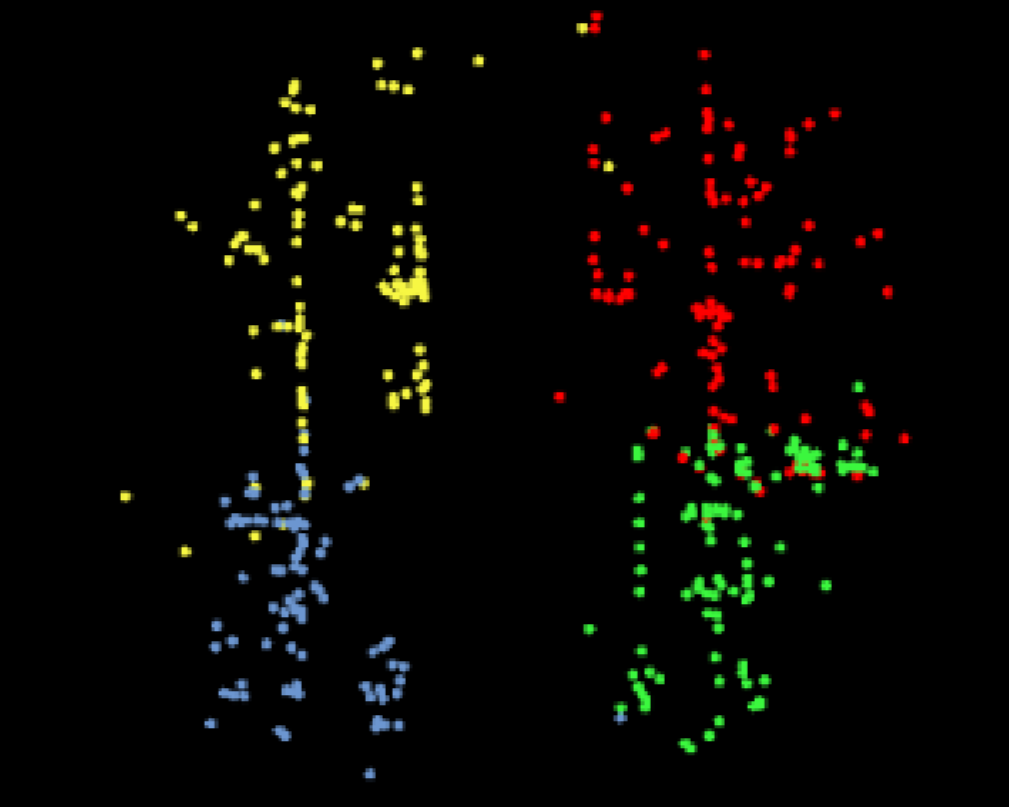}
        \caption{Four Smartmicro UMRR-96 Type 153.}
        \label{fig:4radars}
    \end{subfigure}
    \begin{subfigure}{0.325\textwidth}
        \centering
        \includegraphics[width=\textwidth]{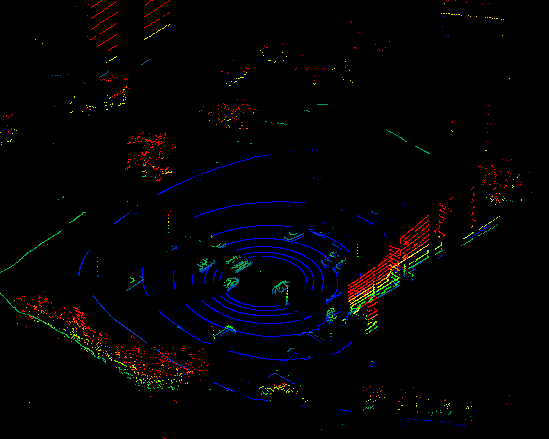}
        \caption{Velodyne VLP-16.}
        \label{fig:velodyne}
    \end{subfigure}
    \begin{subfigure}{0.325\textwidth}
        \centering
        \includegraphics[width=\textwidth]{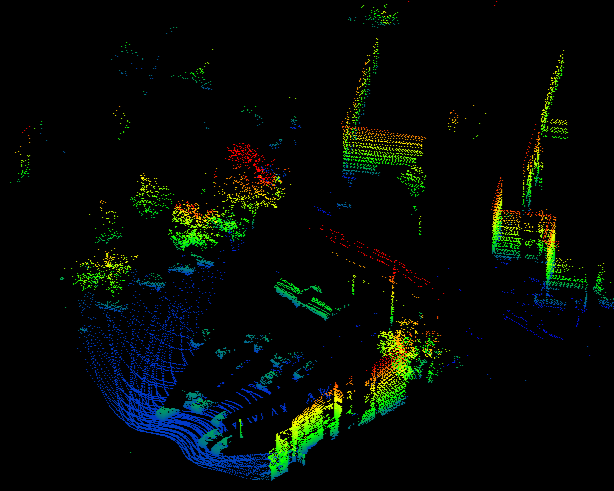}
        \caption{Livox HAP.}
        \label{fig:livox}
    \end{subfigure}

    \caption{Point cloud examples.}
    \label{fig:pointclouds}
\end{figure*}

Another important addition to the dataset is the set of sensors capable of generating point clouds, which can be used for map registration and \ac{slam}. The four \acp{radar} are hardware synchronized, thus generating a unified point cloud with $360^\circ$ coverage, as shown in \figurename~\ref{fig:4radars}. The point clouds generated by the VLP-16 \ac{lidar} and the Livox HAP \ac{lidar} are illustrated in \figurename~\ref{fig:velodyne} and \figurename~\ref{fig:livox}, respectively.

\begin{figure}
    \centering
    \includegraphics[width=0.65\linewidth]{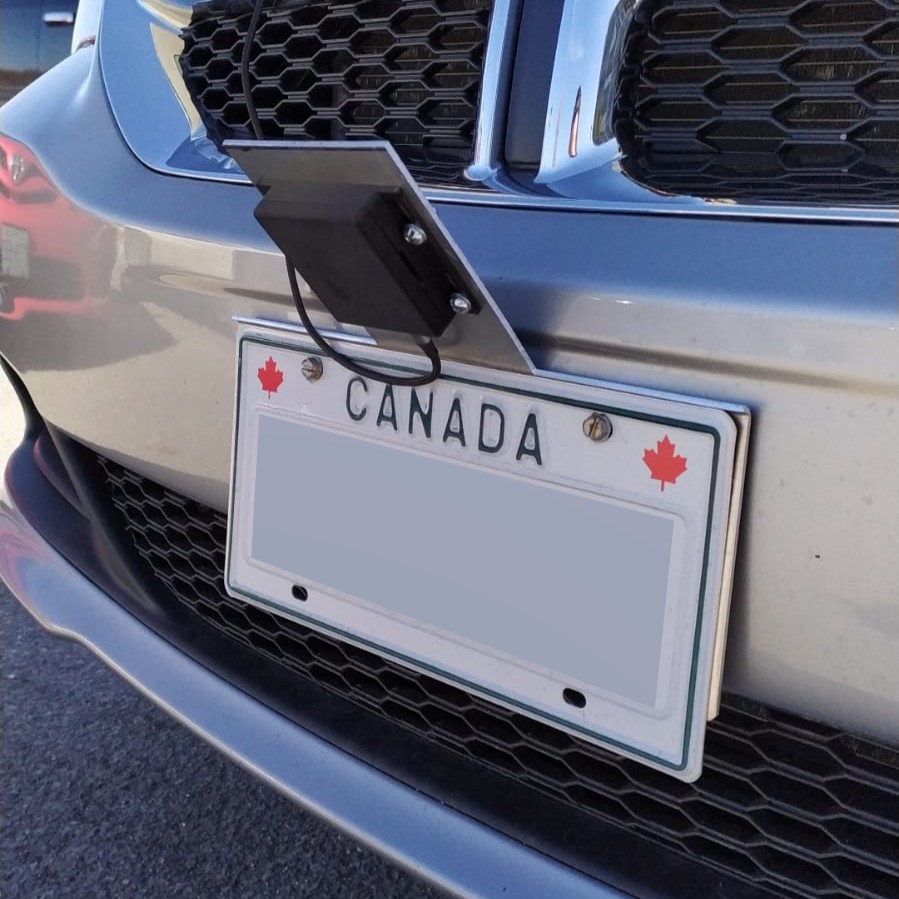}
    \caption{Omnipresense OPS241-A mounted on the front bumper.}
    \label{fig:radar-bumper}
\end{figure}

Finally, the dataset includes a 1D Doppler \ac{radar} mounted on the vehicle's bumper, as presented in \figurename~\ref{fig:radar-bumper}. The sensor is mounted facing the ground at an angle and can only provide the Doppler speed of a single detection. As discussed in section~\ref{sec:forward-speed-radar}, it can be used with an intelligent system to estimate the vehicle's forward speed.

\subsection{Data Collection Framework} \label{sec:trajectories-framework}
\begin{figure*}
\centering
\includegraphics[width= 0.95\linewidth]{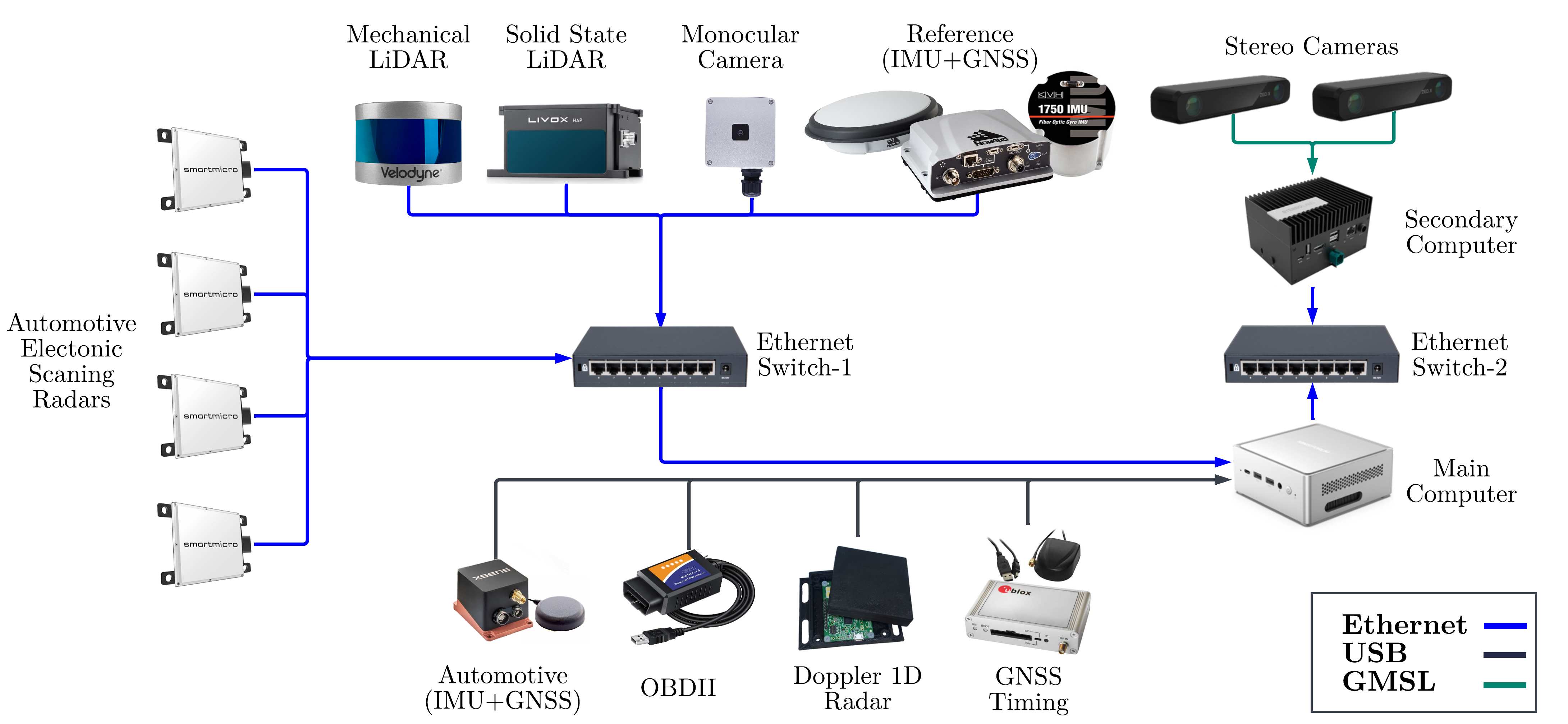}
\caption{The system's architecture including its components and connections.}
\label{fig:system-architectiure}
\end{figure*}

The sensors mentioned above are connected to two computing systems responsible for synchronized data recording: a main computer (13th Gen Core i7, 32G DDR5 RAM, 1TB SSD) and a secondary computer, the NVIDIA Orin™ NX (8-core Arm Cortex, 16 GB LPDDR5 RAM, 1TB SSD). The overall system architecture, including the various sensor connections, is illustrated in \figurename~\ref{fig:system-architectiure}. A range of connection types were employed to accommodate the diverse sensor systems. Some sensors were connected to the main computer via Ethernet-based connectivity through Ethernet \mbox{switch-1}. Other sensors utilize \acf{usb} connectivity through dedicated ports on the main computer. The Stereolabs ZED X cameras, which require high bandwidth, were connected directly to the secondary computer using a \acf{gmsl}. The main and secondary computers are interconnected via Ethernet \mbox{switch-2}, establishing a unified and integrated system.

Given the complexity of the system and the need for a unified time reference across multiple sensor data, time synchronization is deemed essential. To that end, we adopted the \ac{ros}, a robust solution widely used for managing synchronization across multiple devices in the robotics domain~\cite{quigley2009ros}. Both computers run Linux with \ac{ros} installed. A \ac{ros} network was established with the main computer acting as the \ac{ros} master, coordinating communication across the different \ac{ros} nodes. Each computer independently records data from its connected sensors, with \ac{ros} ensuring that all sensor messages are synchronized with the UNIX time of the respective computer.

We employed \ac{ros} software drivers provided by sensor manufacturers or third parties to operate each sensor and obtain data. These drivers were modified and, in some cases, developed to meet our specific requirements. Our system employed \ac{ros}1 and \ac{ros}2 environments simultaneously, necessitating a bridge to convert \ac{ros}2 topics into \ac{ros}1 topics for a unified data recording process. 

To streamline the data collection process, we developed a distributed software tool for synchronized sensor launch, monitoring, and data acquisition. This tool enabled a systematic launch of the necessary sensor drivers, conducting essential tests to detect successful sensor operation, performing periodic checks to detect any communication failures, and locally recording timestamped data from the desired sensors.

While \ac{ros} ensures synchronization of sensor messages within each computer’s UNIX timestamp, time drift can occur between multiple computers, which will cause different sensors recording on different computers to have a misaligned time frame. To address this, we utilized well-established time synchronization tools, such as \textit{Chrony} and \textit{GPSd} services, to create an efficient and transparent time synchronization background process \cite{martins2024time}. \textit{Chrony}, an effective \ac{ntp} implementation, was used to configure the main computer as the \ac{ntp} server and the secondary computer as the \ac{ntp} client, ensuring that the secondary computer’s clock is continuously updated to match the server’s time. Additionally, \textit{GPSd} was used to monitor a dedicated M8T Ublox \ac{gnss} receiver, providing precise timing information. \textit{Chrony} was configured to calibrate the main computer’s time using the \ac{gnss} data, which was then propagated to the secondary computer, ensuring unified \ac{gnss}-aligned time synchronization across the entire \ac{ros} network.

\subsection{Ground Truth} \label{sec:ground truth}
The reference positioning solution is achieved by employing Novatel's advanced PwrPak7-E1 \ac{gnss} receiver, which is seamlessly integrated with the high-end tactical-grade KVH1750 \ac{imu}. To augment the accuracy of the ground truth solutions, stationary antennas were used to conduct differential \ac{gnss} processing, commonly known as \ac{ppk}~\cite{PPK}. Later, the reference solution is post-processed using Inertial Explorer by combining loosely-coupled integration with differential \ac{gnss} along with tightly-coupled integration. 

Positions, namely, latitude, longitude, and altitude are given in the geodetic frame. Velocities
are given with respect to a fixed \ac{enu} frame. The attitude information, namely pitch, roll, and azimuth is given in degrees. Each trajectory includes 50 Hz ground truth data, with an extended version including statistics at 1 Hz. This post-processed information is injected into the \textit{bag} files using \ac{ros} timestamps while keeping the GPS timestamps. Further details about the data format are presented in section~\ref{sec:dataset-format}.

\begin{figure}
    \centering
    \includegraphics[width=0.95\linewidth]{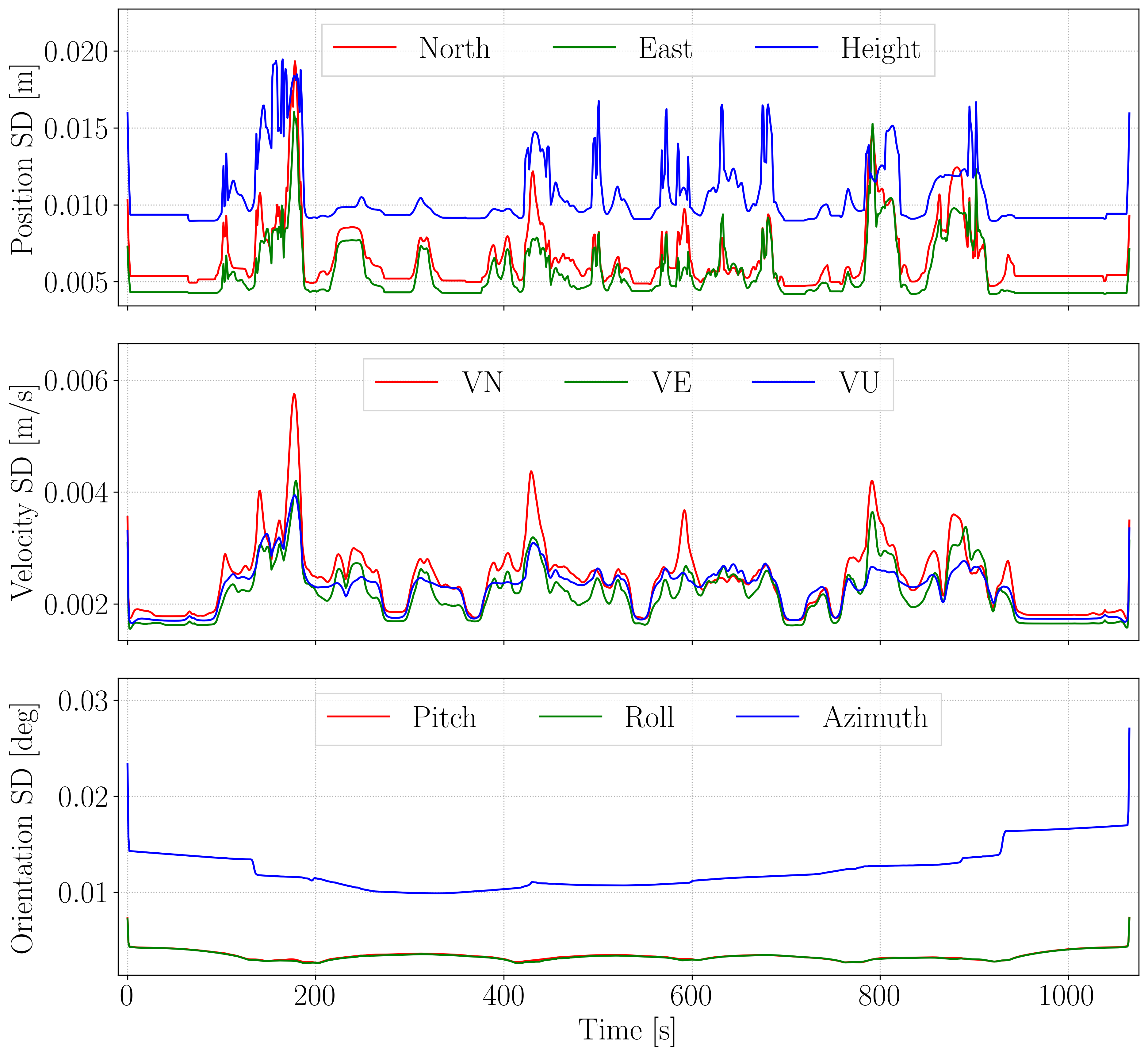}
    \caption{Post-processed standard deviations of position, velocity, and orientation, as reported by Inertial Explorer to trajectory Urban04.}
    \label{fig:SD}
\end{figure}

The standard deviations provided by Inertial Explorer exhibit variation dependent on factors such as satellite visibility (influenced by the environment and time of day), trajectory length, and vehicle dynamics. This information is also available in the \textit{bag} files storing reference data. For position, the average standard deviations fall within the range of 1 centimeter, while for orientation, they are below 0.02$^\circ$ for trajectories \texttt{Urban01} to \texttt{Urban04}. \figurename~\ref{fig:SD} illustrates a sample standard deviations extracted from a trajectory after being post-processed. For trajectories \texttt{Urban05} and \texttt{Urban06}, due to the challenging \ac{gnss} environment, the position standard deviations fall within the range of 50 centimeters, while the orientation standard deviations fall within 0.58$^\circ$ for pitch and roll and 2.08$^\circ$ for heading in some areas.

In indoor environments, where \ac{gnss} signals are significantly degraded, our high-precision \ac{gnss}/\ac{imu} reference system becomes unreliable. As an alternative, we offer a different ground truth solution: the \ac{lmr} approach presented in \cite{mounier2023high}. This method integrates \ac{imu} and vehicle odometer measurements, periodically correcting the vehicle's pose using \ac{lidar} measurements registered to available map data. The \ac{lmr} approach employs an elaborate pipeline that utilizes an extended Kalman filter for sensor fusion and an optimized map filtering algorithm to improve the efficiency of point cloud registration. The \ac{lmr} solution is provided in the local level frame of the map with Cartesian 3D position components in meters and attitude information, i.e., pitch, roll, and azimuth, in degrees, sampled at 50 Hz.

In urban environments, the \ac{lmr} solution has demonstrated sub-meter accuracy, with a root mean square position error of 20 centimeters. Given this strong performance, it is reasonable to expect that this approach will achieve comparable, if not superior, positioning accuracy in indoor environments, particularly due to slower vehicle dynamics. This makes the \ac{lmr} solution a reliable reference for comparison in challenging indoor settings. However, it is important to note that the accuracy of the \ac{lmr} solution is highly dependent on the quality of the offline maps used.


\subsection{Calibration} \label{sec:calib}

\begin{figure}
    \centering
    \includegraphics[width=\linewidth]{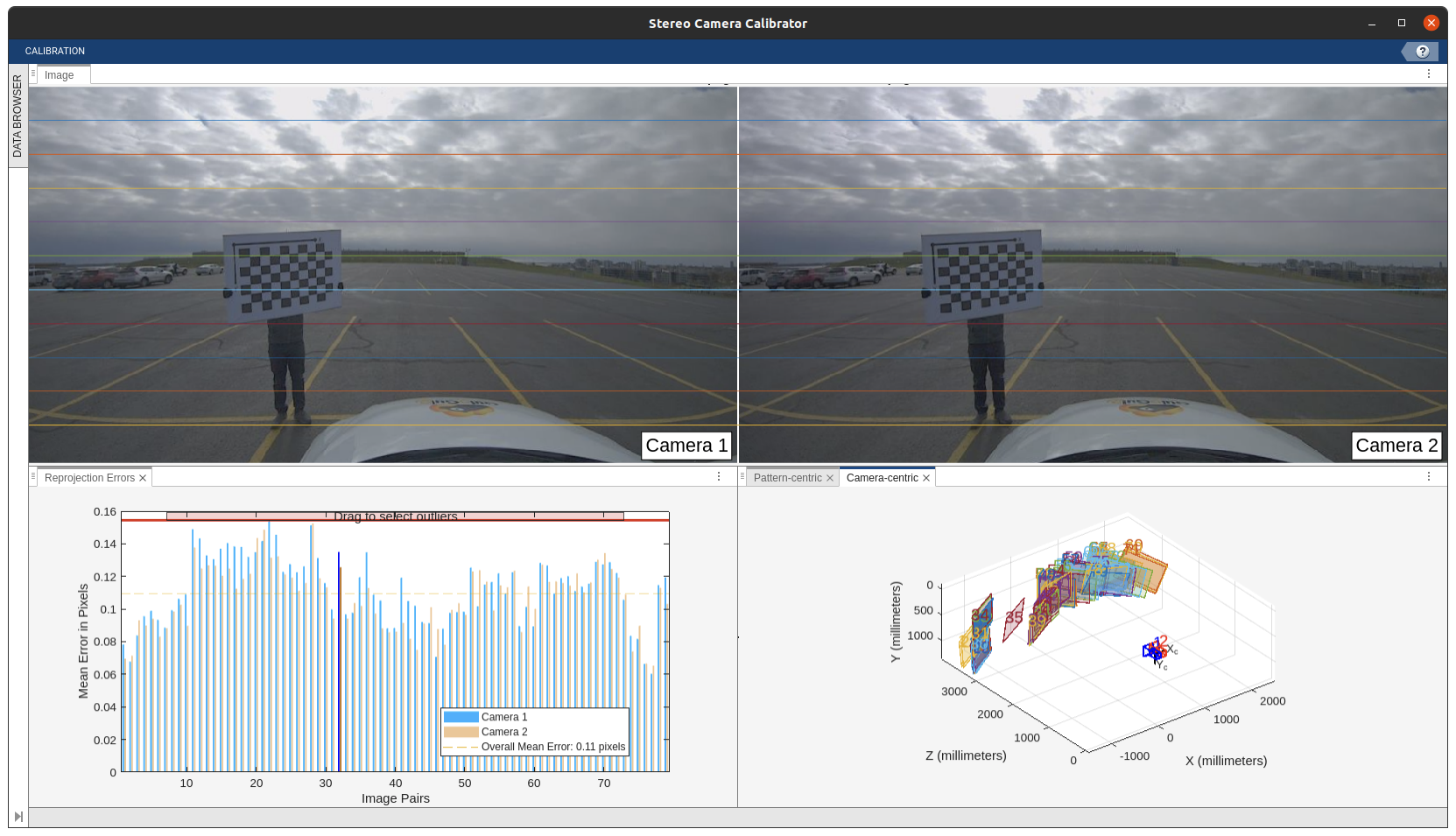}
    \caption{Calibration of the ZED X (left) stereo camera inside MATLAB Stereo Camera Calibrator toolbox.}
    \label{fig:stereo-calib}
\end{figure}

Since our platform includes multiple cameras, we provide some of the calibration parameters used in sections~\ref{sec:mvo} and \ref{sec:svSLAM}. The cameras were calibrated using MATLAB's camera calibrator and stereo camera calibrator toolboxes. To offer greater flexibility to researchers, we also include the \textit{bag} file recorded for calibration, which contains stationary data. This data can be useful for calibrating additional sensors as needed. \figurename~\ref{fig:stereo-calib} illustrates the environment and calibration board used in this \textit{bag} file.

\begin{figure*}
    \centering
    \begin{subfigure}{0.445\textwidth}
        \centering
        \includegraphics[trim={0.65cm 0 0.65cm 0},clip,width=\linewidth]{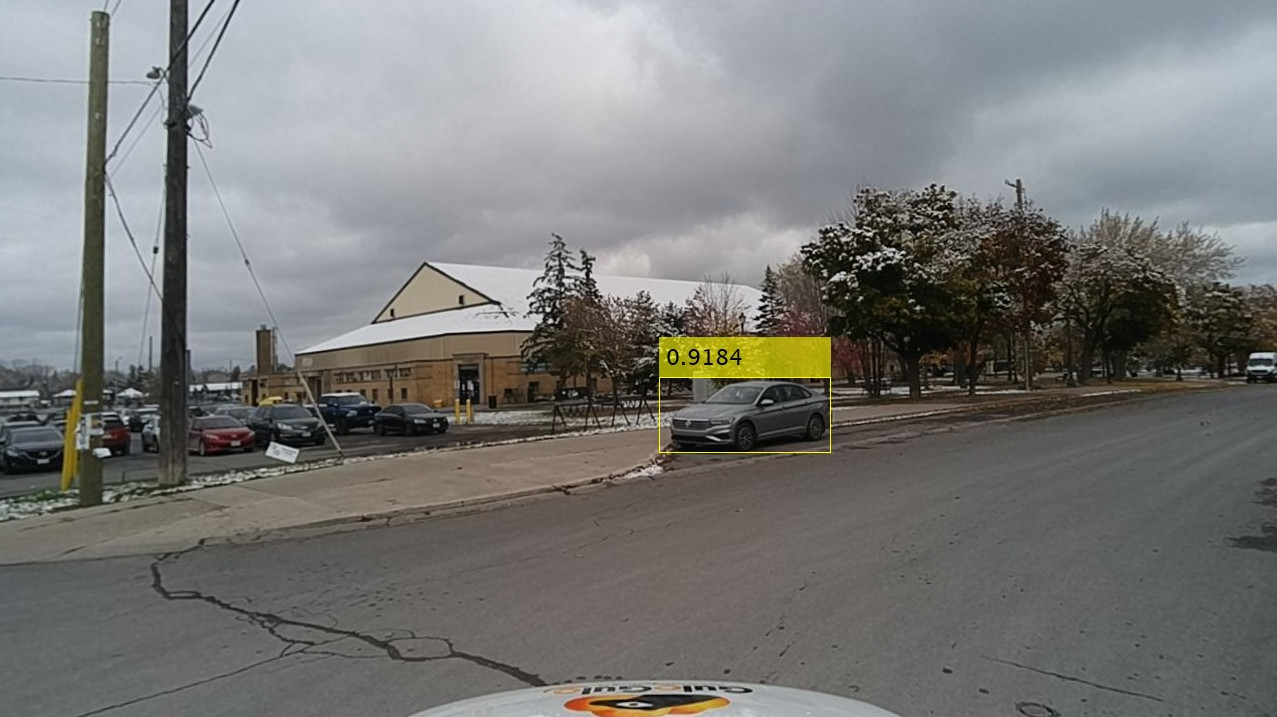}
        \caption{MATLAB Vehicle detector using an OAK camera image.}
        \label{fig:velodyne-oak-detection}
    \end{subfigure}
    \hspace{15pt}
    \begin{subfigure}{0.445\textwidth}
        \centering
        \includegraphics[width=\linewidth]{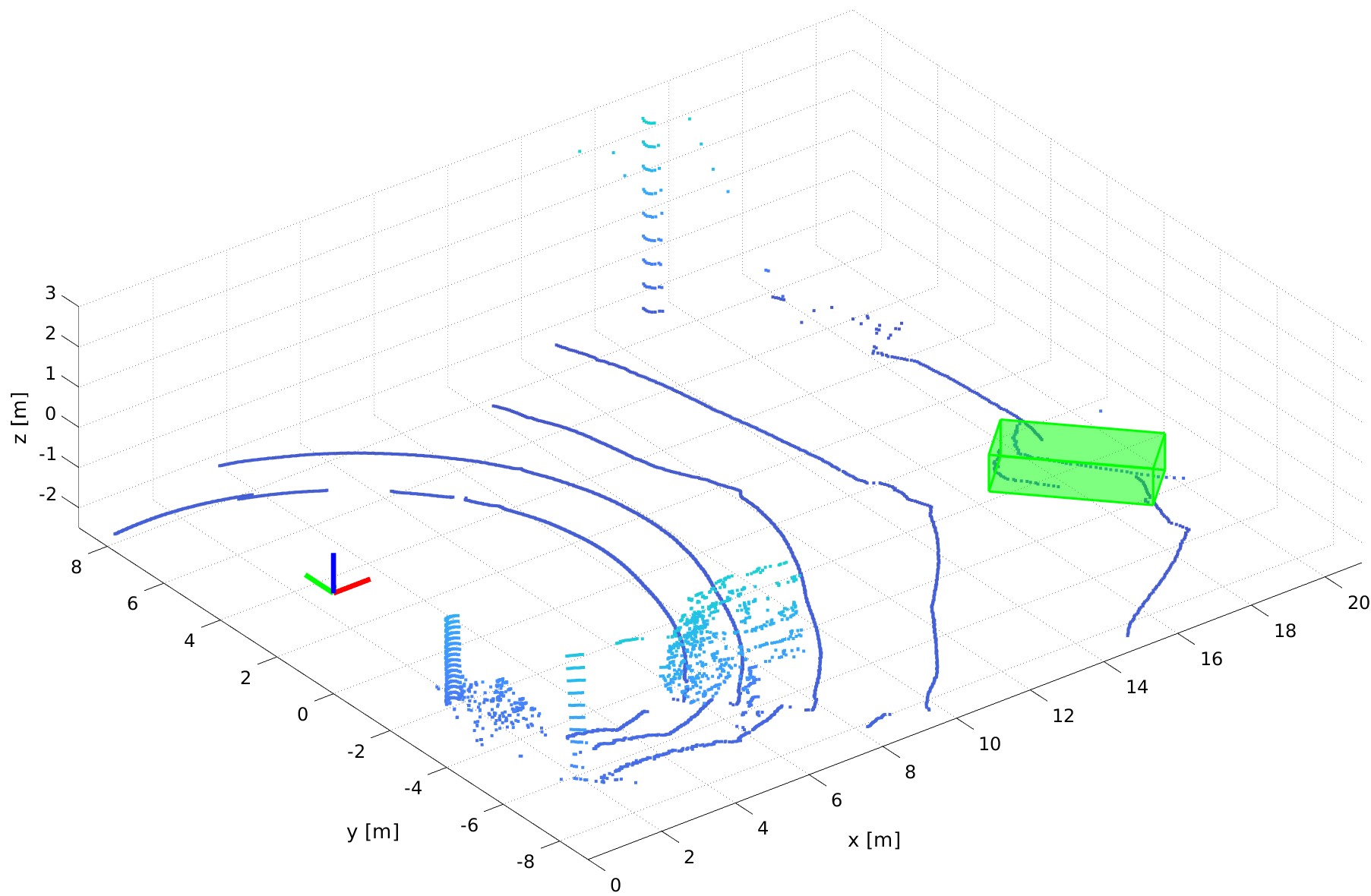}
        \caption{Image bounding box projection onto VLP-16 point cloud.}
        \label{fig:velodyne-oak-bbox}
    \end{subfigure}
    \caption{Camera and \ac{lidar} calibration example.}
    \label{fig:velodyne-oak-calib}
\end{figure*}

We also provide the calibration between the OAK camera and the VLP-16 \ac{lidar}. These sensors were calibrated using MATLAB's \ac{lidar} Camera Calibrator toolbox and the calibration \textit{bag}. \figurename~\ref{fig:velodyne-oak-calib} shows an example of the estimated calibration parameters. In this example, a vehicle is detected using the MATLAB vehicle detector, and the bounding box of the detection is then projected onto the VLP-16 point cloud.

The reference system undergoes a calibration routine, following Novatel's recommendations, to calculate the Body-to-Vehicle Frame Rotation, ensuring accurate kinematic alignment. This alignment has already been compensated for in the provided solutions. The remaining sensors were carefully assembled according to the 3D CAD system (\figurename~\ref{fig:coordsys}), with all distances measured to an uncertainty of less than 1 centimeter. We also provide \ac{ros} launch files that include the transformation tree connecting all sensors.

\section{Trajectories} \label{sec:trajectories}

\begin{figure*}
    \centering
    \begin{subfigure}{0.325\textwidth}
        \centering
        \includegraphics[trim={4cm 0 4cm 0},clip,width=\textwidth]{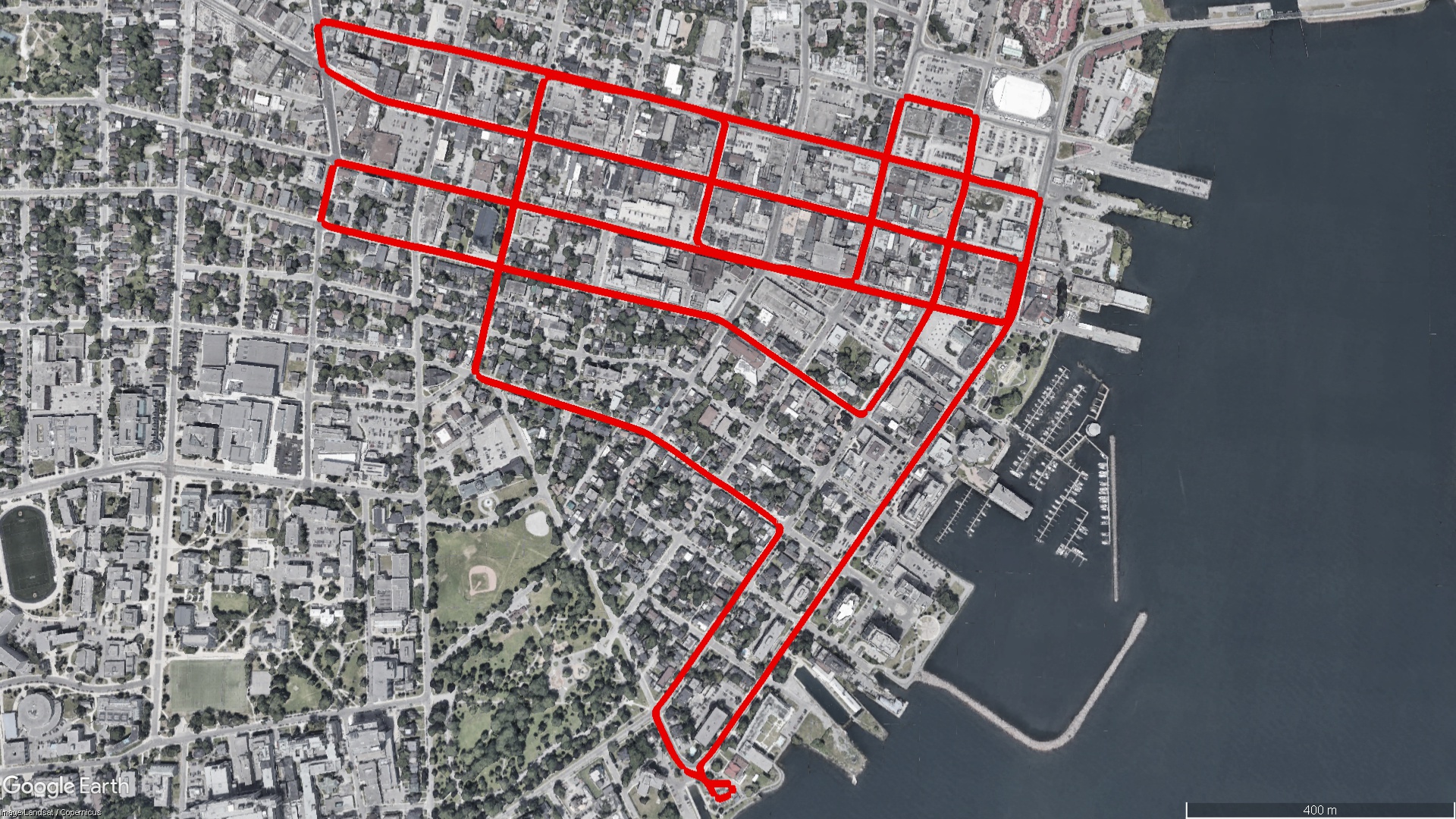}
        \caption{\texttt{Urban01}.}
        \label{fig:urban01}
    \end{subfigure}
    \begin{subfigure}{0.325\textwidth}
        \centering
        \includegraphics[trim={4cm 0 4cm 0},clip,width=\textwidth]{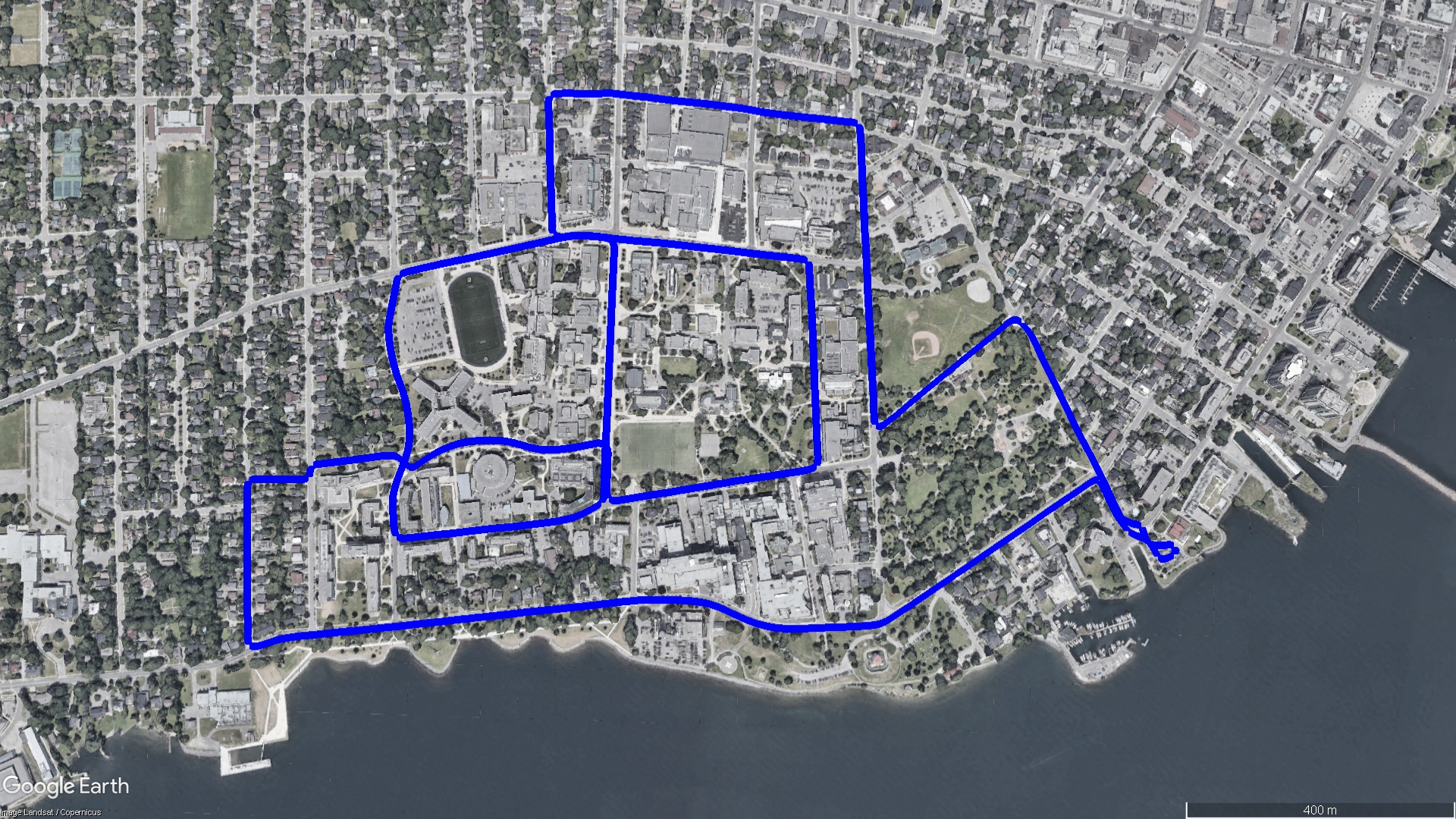}
        \caption{\texttt{Urban02}.}
        \label{fig:urban02}
    \end{subfigure}
    \begin{subfigure}{0.325\textwidth}
        \centering
        \includegraphics[trim={4cm 0 4cm 0},clip,width=\textwidth]{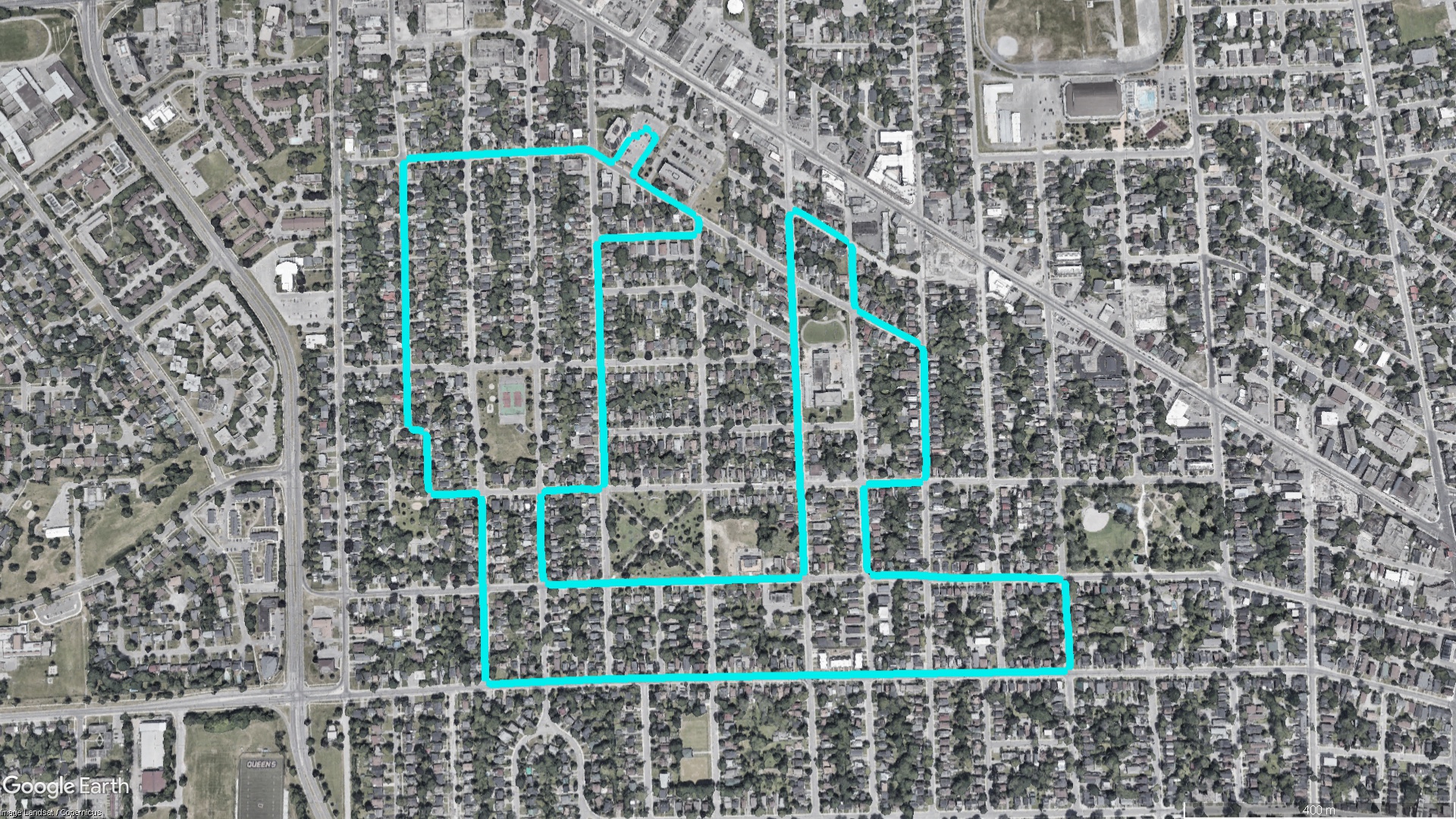}
        \caption{\texttt{Urban03}.}
        \label{fig:urban03}
    \end{subfigure} \\ \vspace{12pt}
    \begin{subfigure}{0.325\textwidth}
        \centering
        \includegraphics[trim={4cm 0 4cm 0},clip,width=\textwidth]{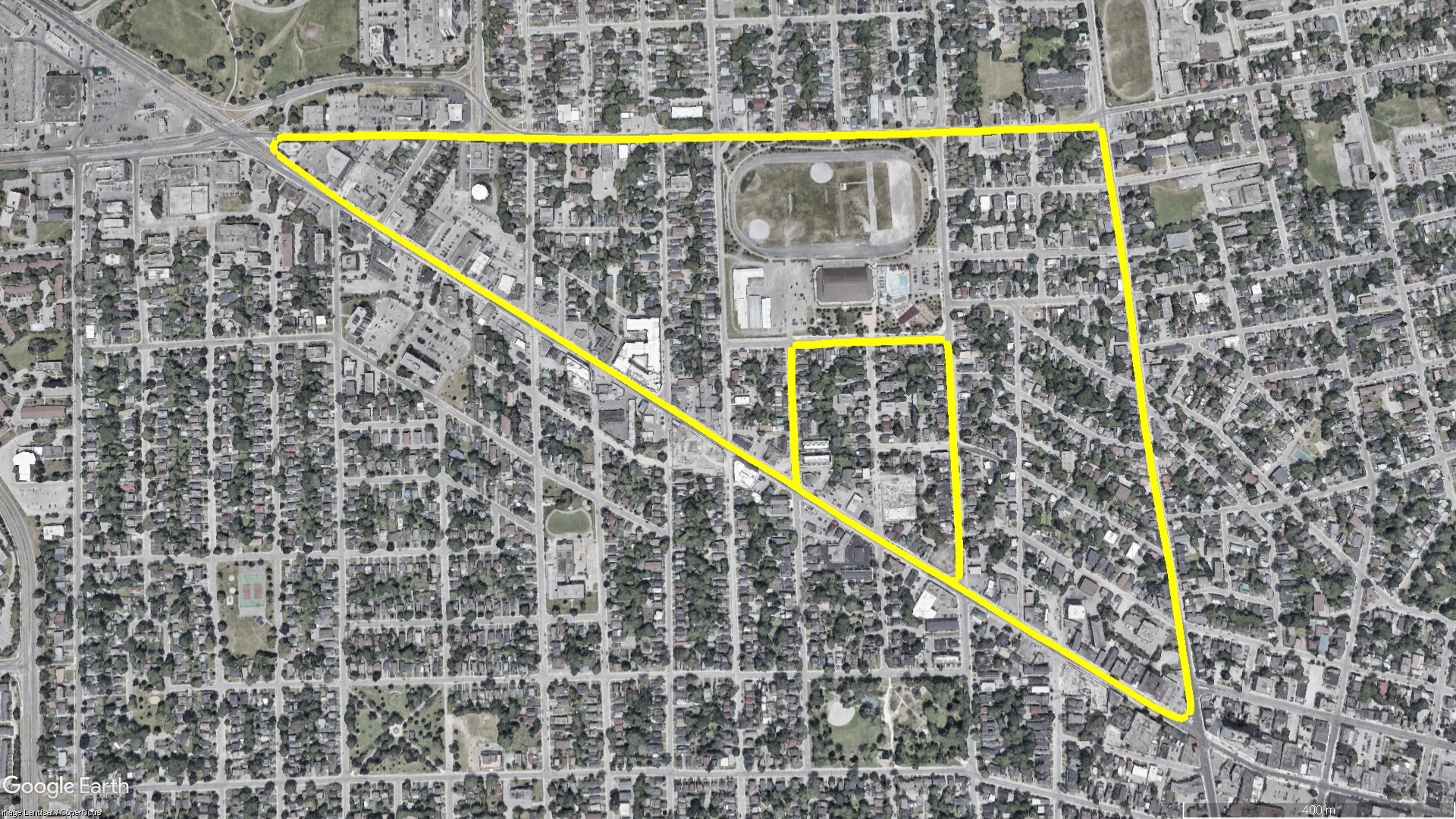}
        \caption{\texttt{Urban04}.}
        \label{fig:urban04}
    \end{subfigure}
    \begin{subfigure}{0.325\textwidth}
        \centering
        \includegraphics[trim={4cm 0 4cm 0},clip,width=\textwidth]{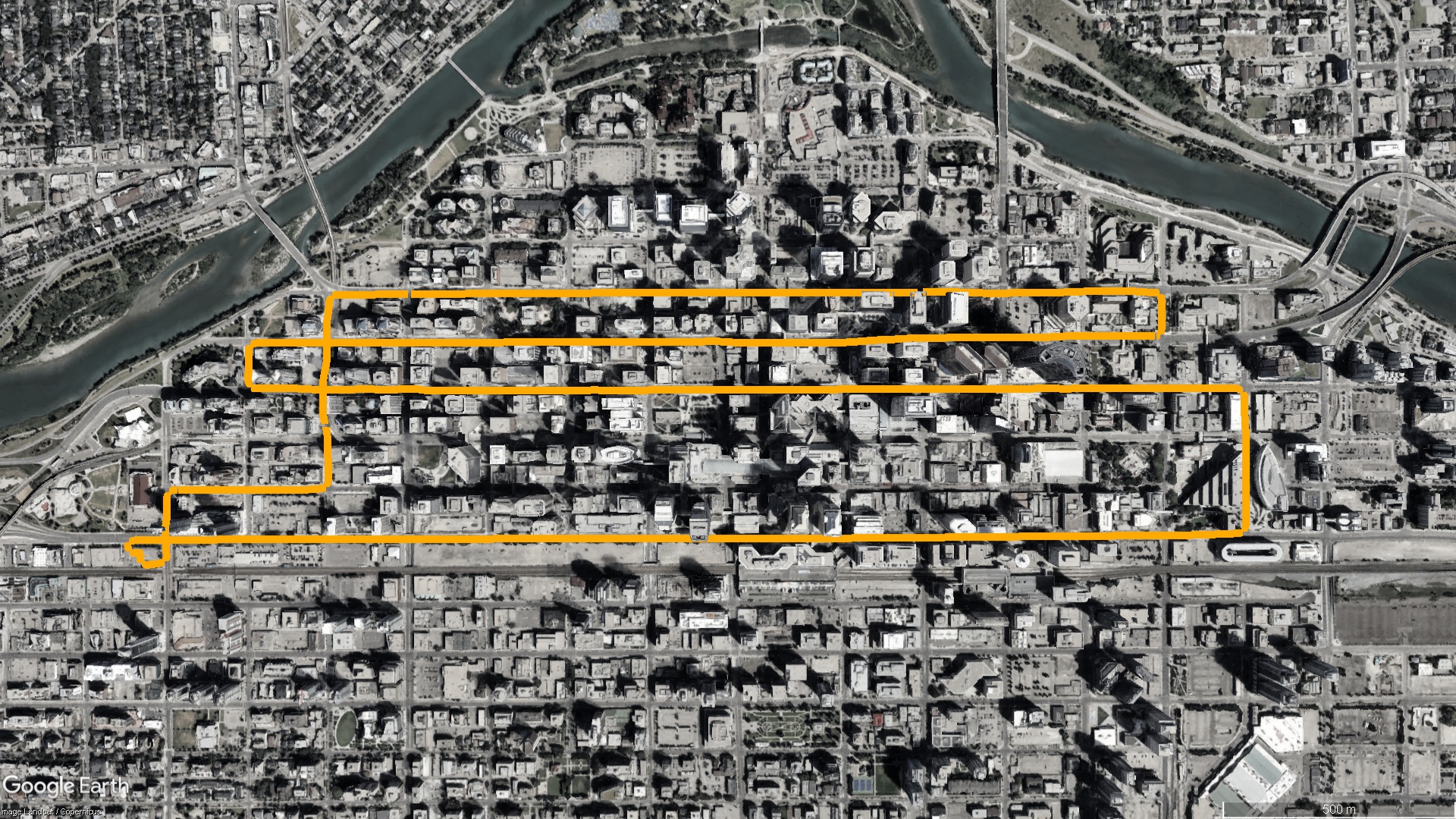}
        \caption{\texttt{Urban05}.}
        \label{fig:urban05}
    \end{subfigure}
    \begin{subfigure}{0.325\textwidth}
        \centering
        \includegraphics[trim={4cm 0 4cm 0},clip,width=\textwidth]{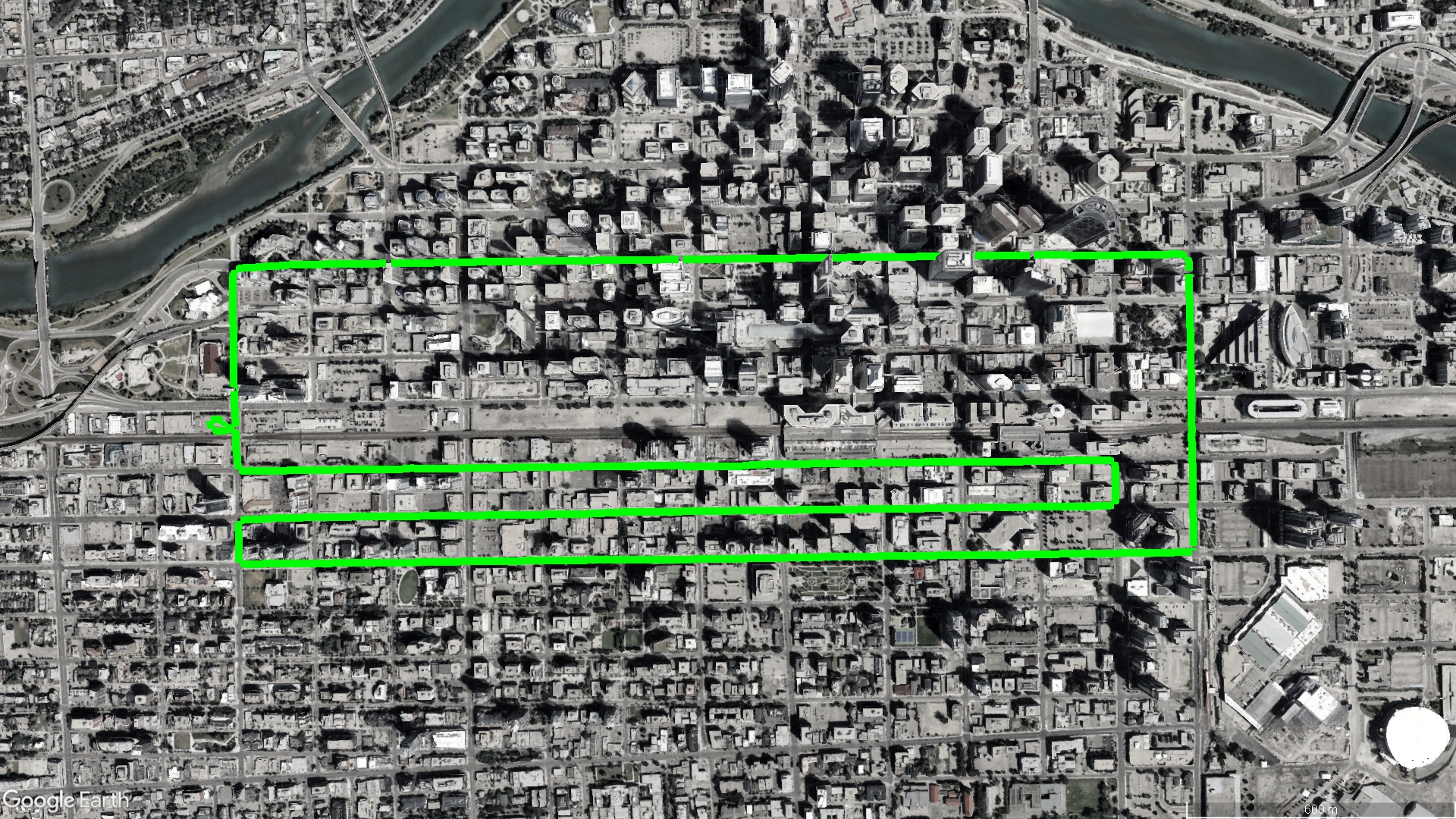}
        \caption{\texttt{Urban06}.}
        \label{fig:urban06}
    \end{subfigure}
    \caption{Outdoor trajectories.}
    \label{fig:outdoor-trajectories}
\end{figure*}

The dataset includes both outdoor and indoor trajectory data recorded in urban environments across Kingston, ON, and Calgary, AB, in Canada. Kingston is primarily a residential area with abundant vegetation, while downtown Calgary is a dense urban setting with numerous high-rise buildings. These two environments present distinct challenges for autonomous vehicle navigation. \figurename~\ref{fig:outdoor-trajectories} shows the six outdoor trajectories and their metadata is detailed in \tablename~\ref{tab:metadata}.


\begin{table}
\renewcommand{\arraystretch}{0.8}
\centering
\caption{Metadata of the trajectories. (D) for a daytime trajectory and (N) for a nighttime trajectory.}
\label{tab:metadata}
\begin{tabular}{@{}lcccc@{}}
\toprule
\textbf{Trajectory} & \makecell[c]{\textbf{Length}\\{(km)}} & \makecell[c]{\textbf{Duration}\\{(min)}} & \makecell[c]{\textbf{Avg. Speed}\\{(km/h)}}  & \makecell[c]{\textbf{Max. Speed}\\{(km/h)}}\\ \midrule
\texttt{Urban01 (D)}  & 8.70 & 35.42 & 14.75 & 47.11\\
\texttt{Urban01 (N)}  & 8.68 & 31.12 & 16.74 & 44.87\\
\texttt{Urban02 (D)}  & 6.78 & 23.58 & 17.26 & 40.49\\
\texttt{Urban02 (N)}  & 6.79 & 24.20 & 16.82 & 40.61\\
\texttt{Urban03 (D)}  & 5.21 & 21.53 & 14.51 & 43.05\\
\texttt{Urban03 (N)}  & 5.17 & 18.50 & 16.77 & 45.43\\
\texttt{Urban04 (D)}  & 4.87 & 17.75 & 16.46 & 44.57\\
\texttt{Urban04 (N)}  & 4.87 & 16.42 & 17.79 & 51.85\\
\texttt{Urban05 (D)}  & 9.77 & 34.25 & 17.12 & 44.49\\
\texttt{Urban06 (D)}  & 9.61 & 36.55 & 15.78 & 44.37\\
\texttt{Indoor01} & 1.25 & 7.03 & 10.63 & 23.52 \\
\texttt{Indoor02} & 0.97 & 5.86 & 9.93  & 22.13 \\ 
\texttt{Indoor03} & 1.27 & 8.43 & 9.00  & 22.20 \\
\texttt{Indoor04} & 1.80 & 14.20 & 7.48  & 12.69 \\
\texttt{Indoor05} & 1.25 & 9.90 & 7.58  & 15.17 \\ \midrule
\textbf{Total} & 76.99 & 304.74 & - & - \\ \bottomrule
\end{tabular}
\end{table}

All trajectories followed the same protocol. The reference system was initialized in an open-sky area. Next, all sensors were powered up to ensure that no biases accumulated during the initialization of the reference system. The main computer was then time-synchronized using the \ac{gnss} timing receiver. Subsequently, the secondary computer was synchronized to the main computer, and recording was initiated. The vehicle remained stationary at the beginning and end of the trajectories for approximately 2 minutes.

\begin{figure*}
    \centering
    \begin{subfigure}{0.325\textwidth}
        \centering
        \includegraphics[width=\textwidth]{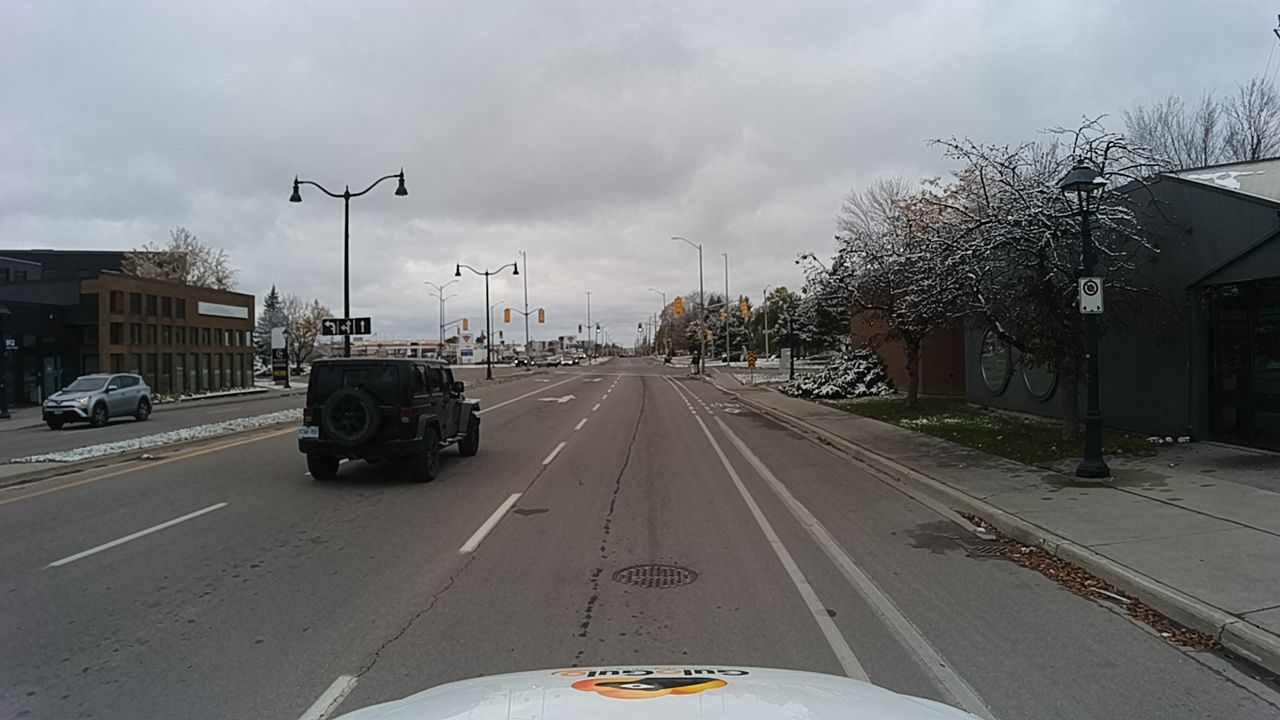}
        \caption{Day.}
        \label{fig:oak-day}
    \end{subfigure}
    \begin{subfigure}{0.325\textwidth}
        \centering
        \includegraphics[width=\textwidth]{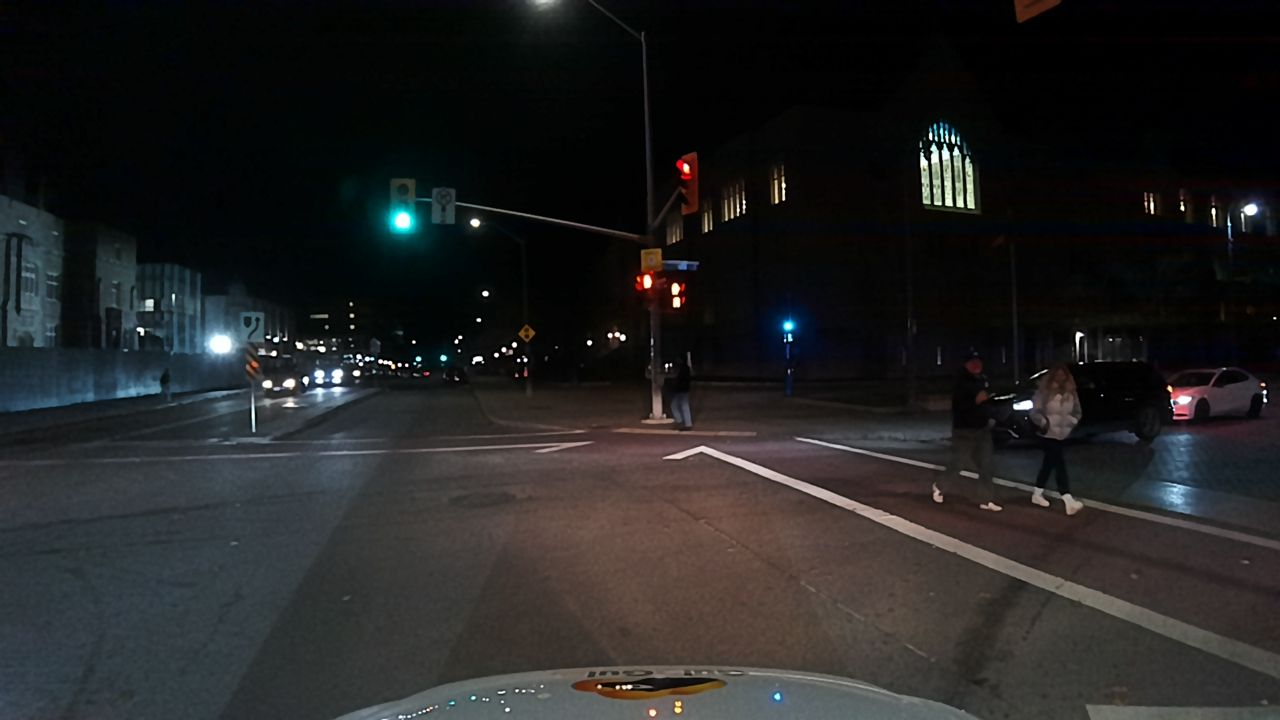}
        \caption{Night.}
        \label{fig:oak-night}
    \end{subfigure}
    \begin{subfigure}{0.325\textwidth}
        \centering
        \includegraphics[width=\textwidth]{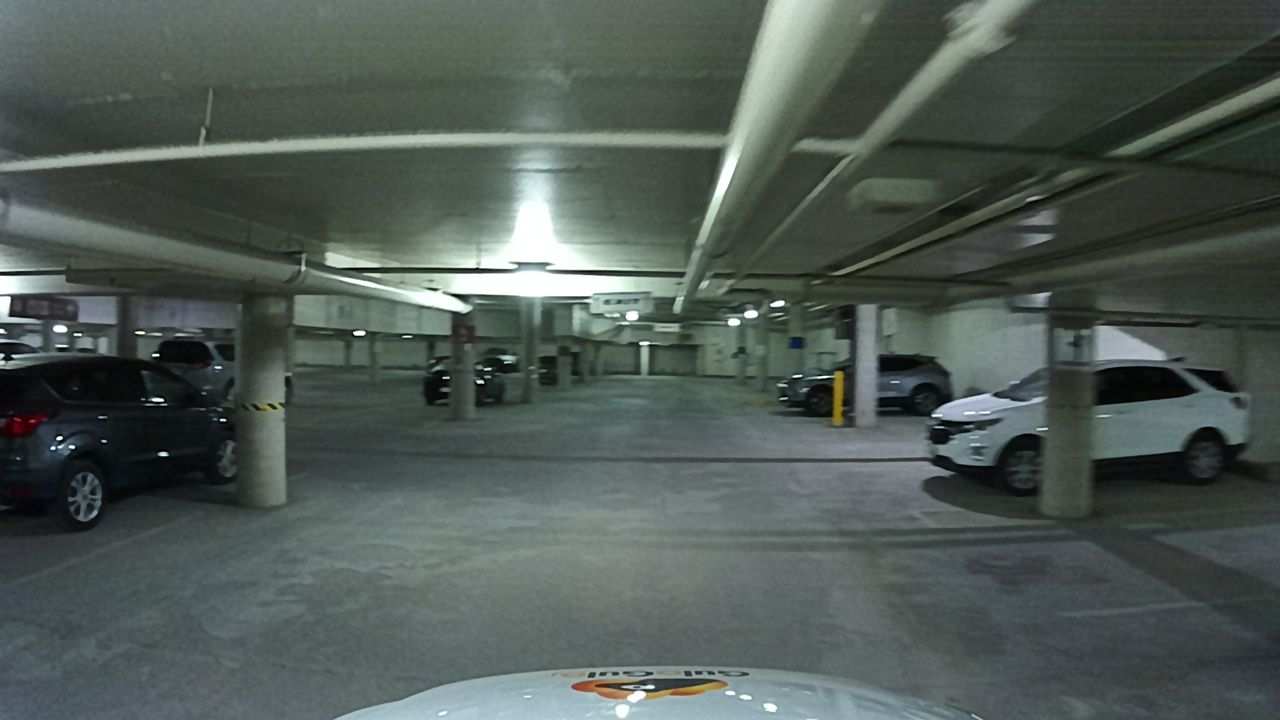}
        \caption{Indoor.}
        \label{fig:oak-indoor}
    \end{subfigure}
    \caption{OAK camera images under different lighting conditions.}
    \label{fig:oak-camera-conditions}
\end{figure*}

The outdoor trajectories were collected during both day and night to enable the investigation of computer vision-based methods for urban navigation under different lighting conditions, see \figurename~\ref{fig:oak-camera-conditions}. Therefore, eight outdoor trajectories are provided, totaling approximately 50 kilometers of recorded data.

\begin{figure*}
    \centering
    \begin{subfigure}{0.325\textwidth}
        \centering
        \includegraphics[trim={4cm 0 4cm 0},clip,width=\textwidth]{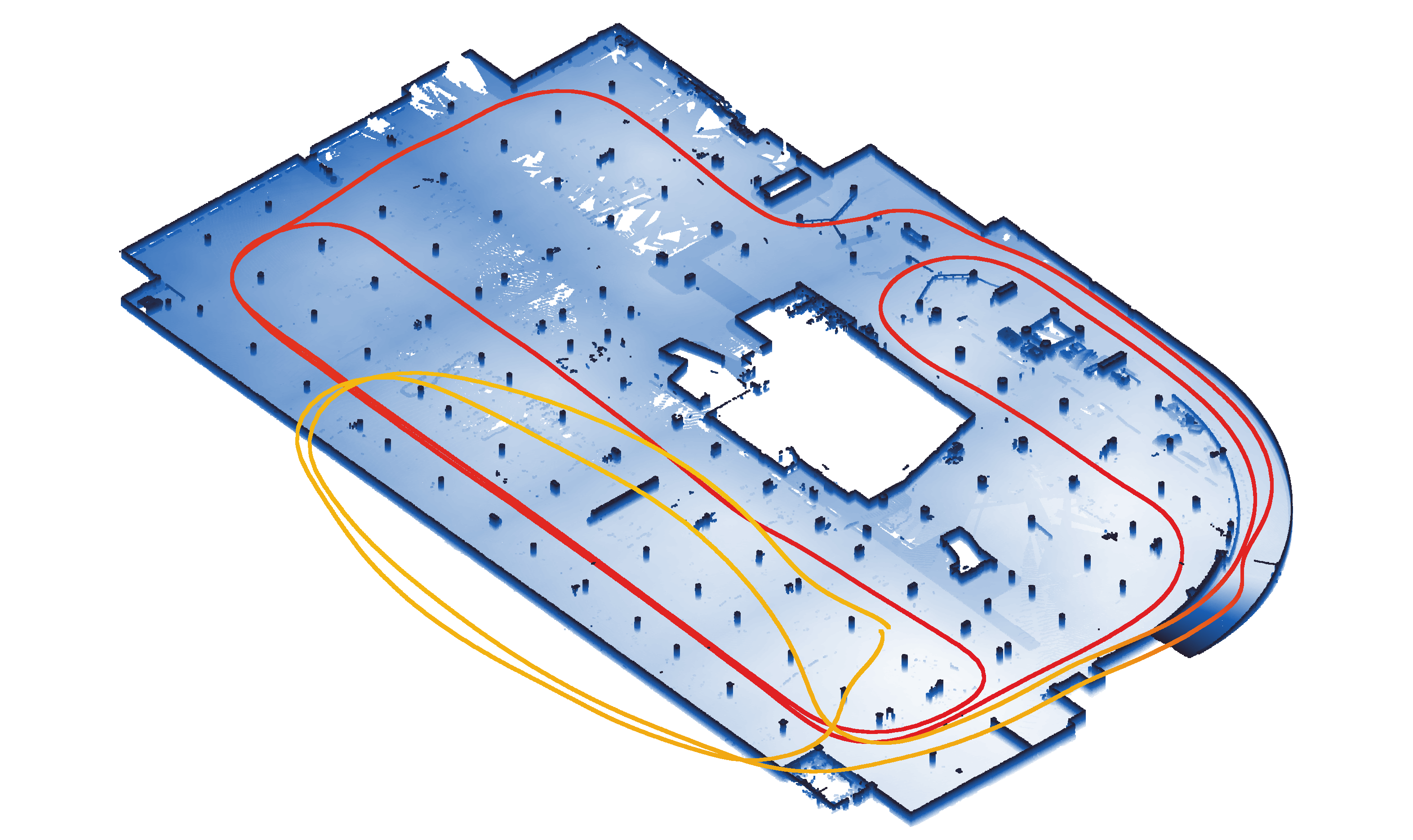}
        \caption{\texttt{Indoor01}.}
        \label{fig:indoor01}
    \end{subfigure}
    \begin{subfigure}{0.325\textwidth}
        \centering
        \includegraphics[trim={4cm 0 4cm 0},clip,width=\textwidth]{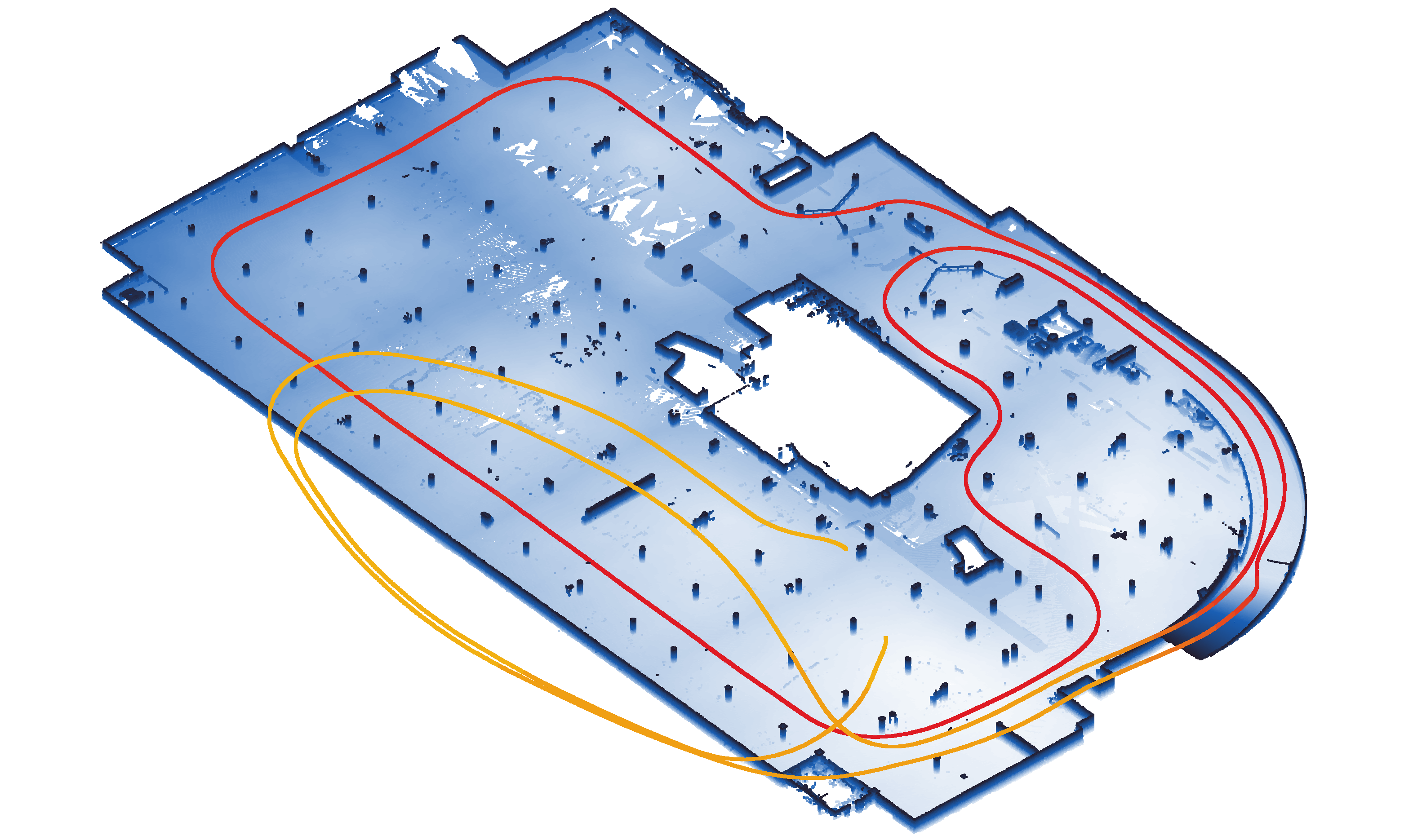}
        \caption{\texttt{Indoor02}.}
        \label{fig:indoor02}
    \end{subfigure}
    \begin{subfigure}{0.325\textwidth}
        \centering
        \includegraphics[trim={4cm 0 4cm 0},clip,width=\textwidth]{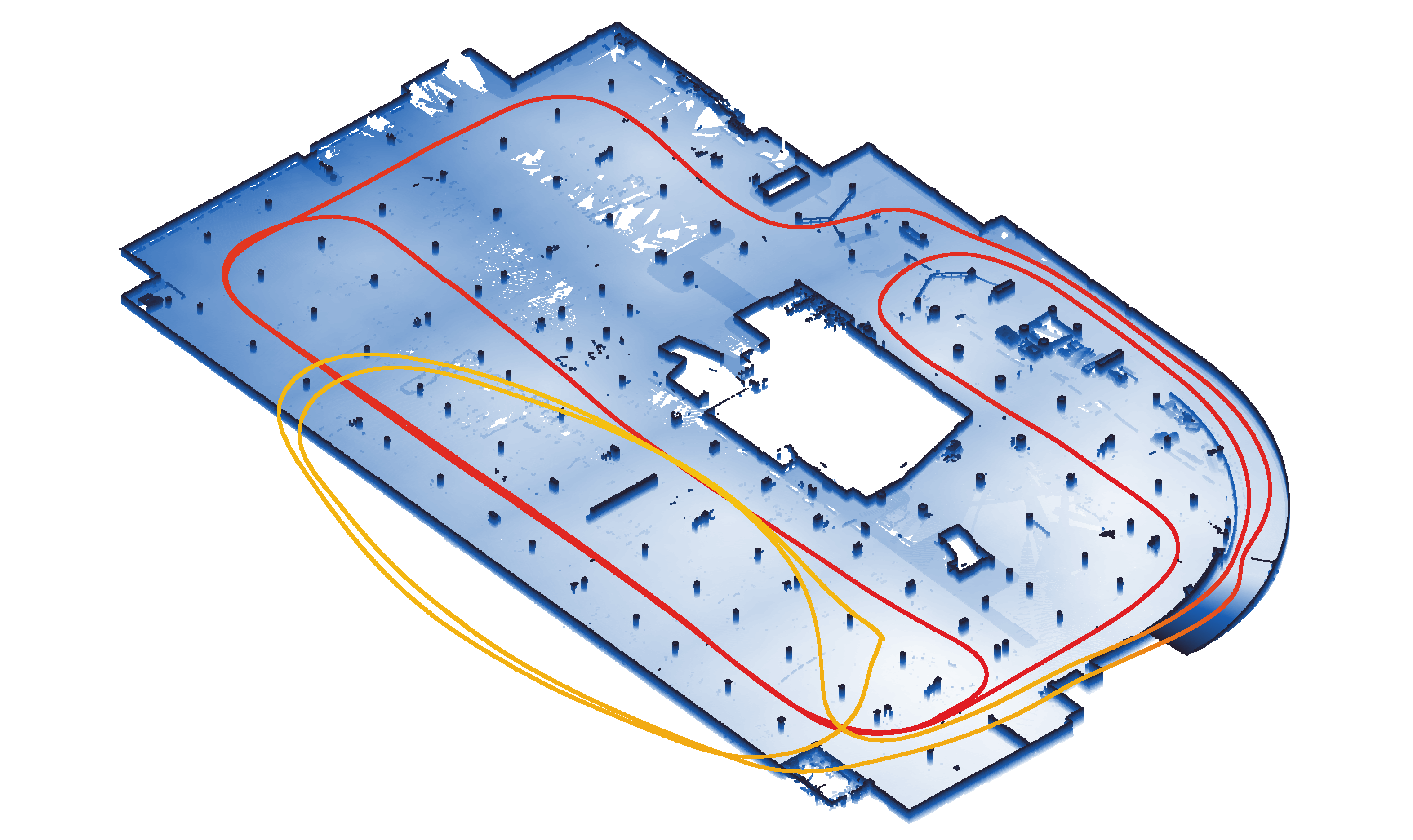}
        \caption{\texttt{Indoor03}.}
        \label{fig:indoor03}
    \end{subfigure} \\ \vspace{12pt}
    \begin{subfigure}{0.425\textwidth}
        \centering
        \includegraphics[trim={6cm 0 6cm 1cm},clip,width=\textwidth]{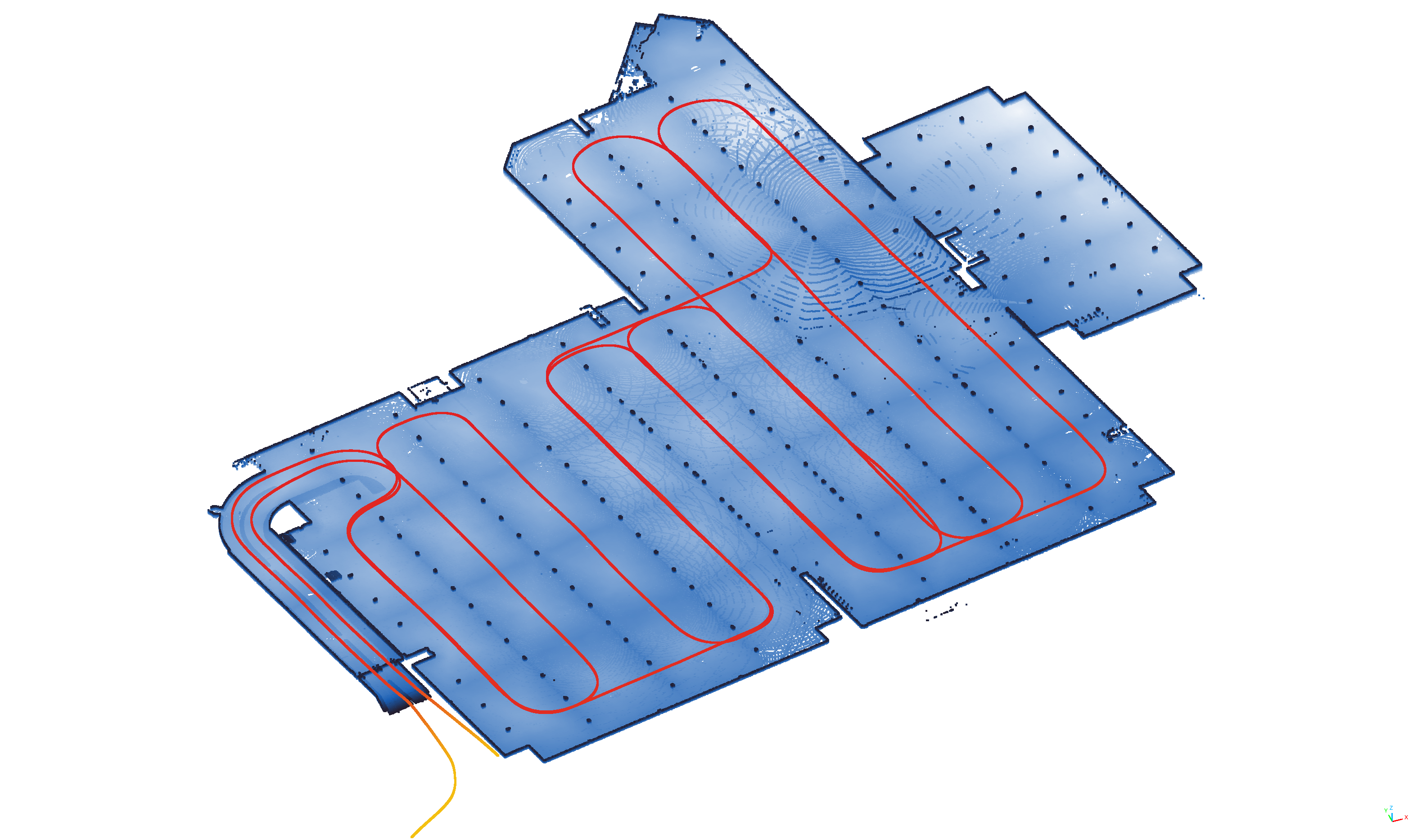}
        \caption{\texttt{Indoor04}.}
        \label{fig:indoor04}
    \end{subfigure}
    \begin{subfigure}{0.425\textwidth}
        \centering
        \includegraphics[trim={6cm 0 6cm 1cm},clip,width=\textwidth]{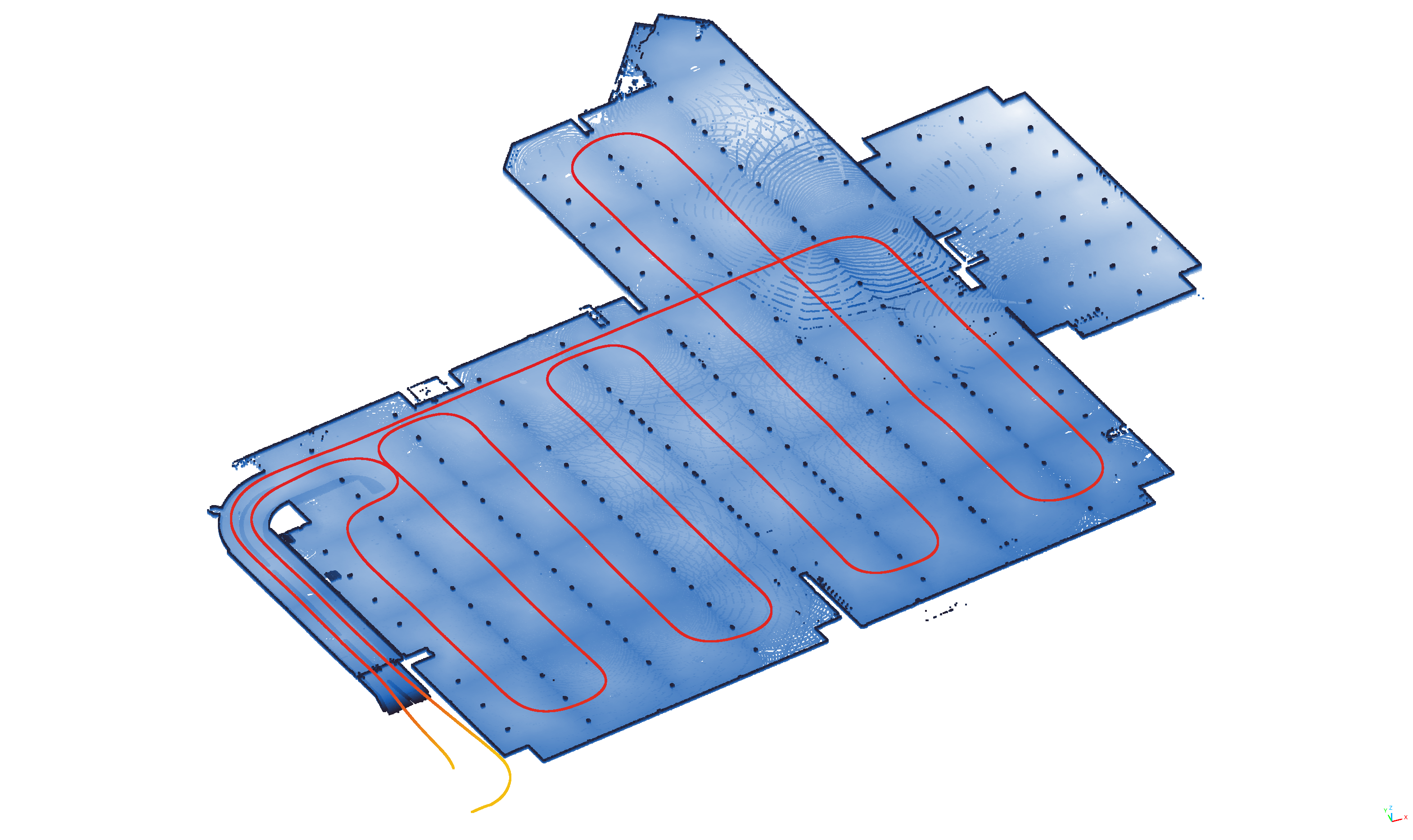}
        \caption{\texttt{Indoor05}.}
        \label{fig:indoor05}
    \end{subfigure}
    \caption{Indoor trajectories. Segments are colored according to their height, with the \textcolor{YellowOrange}{\textbf{yellow}} portion recorded outside the garage and the \textcolor{red}{\textbf{red}} portion recorded inside. The garage ceiling is rendered transparent to enhance visualization.}
    \label{fig:indoor-trajectories}
\end{figure*}

The dataset includes five indoor trajectories recorded in two indoor underground garages, as illustrated in \figurename~\ref{fig:indoor-trajectories}. As shown in \figurename~\ref{fig:oak-camera-conditions}(c), the selected garages are confined spaces that pose multiple challenges for perception sensors. To further contribute to the community, we provide high-definition 3D maps of both garages, which were prepared using a high-definition stationary \ac{lidar}. All trajectories started outside where \ac{gnss} was available for proper initialization of the respective reference system. After a few minutes, the vehicle was driven throughout the garage before returning outside to complete the recording.

\section{Dataset Organization} \label{sec:dataset-format}
\begin{figure}
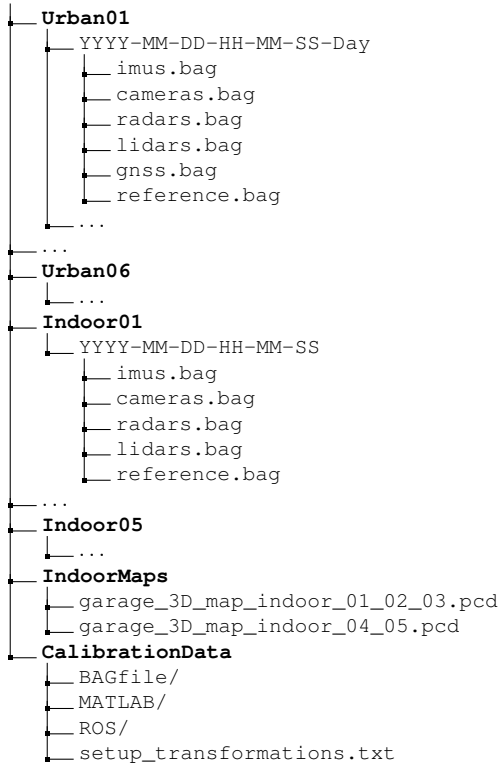

\begin{footnotesize}
{
\renewcommand{\DTbaselineskip}{1.2em}
\dirtree{%
    .1 .
        .2 \textbf{Urban01}.
            .3 YYYY-MM-DD-HH-MM-SS-Day.
                .4 imus.bag.
                .4 cameras.bag.
                .4 radars.bag.
                .4 lidars.bag.
                .4 gnss.bag.
                .4 reference.bag.
            .3 $\dots$.
        .2 $\dots$.
        .2 \textbf{Urban06}.
            .3 $\dots$.
        .2 \textbf{Indoor01}.
            .3 YYYY-MM-DD-HH-MM-SS.
                .4 imus.bag.
                .4 cameras.bag.
                .4 radars.bag.
                .4 lidars.bag.
                .4 reference.bag.
        .2  $\dots$.
        .2 \textbf{Indoor05}.
            .3 $\dots$.
        .2 \textbf{IndoorMaps}.
            .3 garage\_3D\_map\_indoor\_01\_02\_03.pcd.
            .3 garage\_3D\_map\_indoor\_04\_05.pcd.
        .2 \textbf{CalibrationData}.
            .3 BAGfile/.
            .3 MATLAB/.
            .3 ROS/.
            .3 setup\_transformations.txt.
}
}
\end{footnotesize}
\caption{Data organization.}
\label{fig:folder-structure}
\end{figure}

The dataset is organized as shown in \figurename~\ref{fig:folder-structure}. Both day and night outdoor trajectories are organized into folders named as $\texttt{Urban01}\dots\texttt{Urban06}$. Inside these subfolders named with a specific timestamp, one can find various \textit{bag} files named according to their contents. For example, all camera-related data are stored in the \texttt{cameras.bag} file. We structured the \textit{bag} files this way to facilitate their use; users can download only the \textit{bag} files of interest and play them together using the command \texttt{rosbag play *}, assuming all \textit{bag} files are stored in the same directory. In addition to the trajectory folders, the user can find the \textit{IndoorMaps} folder, which contains the PCD format files of the indoor maps, and the \textit{CalibrationData} folder, which includes the calibration files and MATLAB projects of the tools discussed in section~\ref{sec:calib}, the \textit{bag} file recorded for calibration, the ROS launch file to launch the transformation tree and the transformations considering KVH1750 as the origin (see file \texttt{setup\_transformation.txt}).

\section{Demonstrations} \label{sec:demonstrations}
To demonstrate our dataset's validity, we test various navigation and positioning algorithms. The objective of the demonstrations is to validate the functionality of the data, rather than proposing, validating, or criticizing positioning methods.

\subsection{Evaluation Metrics}
To evaluate the performance of the tested systems while using the \ac{navinst} dataset, we use two error metrics:
\begin{enumerate}
    \item \ac{ate}: Measures the root-mean-square error between the predicted 3D pose and the ground truth to assess the global consistency of the estimated trajectory. We split the analysis into translational error (\textit{trans}) expressed in meters and rotational error (\textit{rot}) expressed in degrees.
    \item \ac{rpe}: Measures the error in the relative motion between consecutive poses, which helps in assessing the drift in the estimated trajectory. It is computed as the root-mean-square error between the estimated and the actual relative motions over a fixed time interval. We split the analysis into translational error (\textit{trans}) expressed in meters and rotational error (\textit{rot}) expressed in degrees.
\end{enumerate}

\subsection{LiDAR Odometry}
For \acp{lidar}, we demonstrate the data validity using \ac{lo} with the renowned Kiss-ICP algorithm \cite{vizzo2023kiss}. This method has proven to deliver efficient and competitive state-of-the-art performance in odometry-based pose estimation, relying on \ac{lidar}-only measurements.

We utilized point cloud data from both \acp{lidar} across two different urban trajectories, resulting in two separate \ac{lo} solutions: Velodyne-\ac{lo} and Livox-\ac{lo}. The obtained solutions in each trajectory were compared to the high-precision positioning reference solution to assess their accuracy.

\begin{figure*}
    \centering
    \begin{subfigure}{0.43\textwidth}
        \centering
        \includegraphics[height=0.8\linewidth]{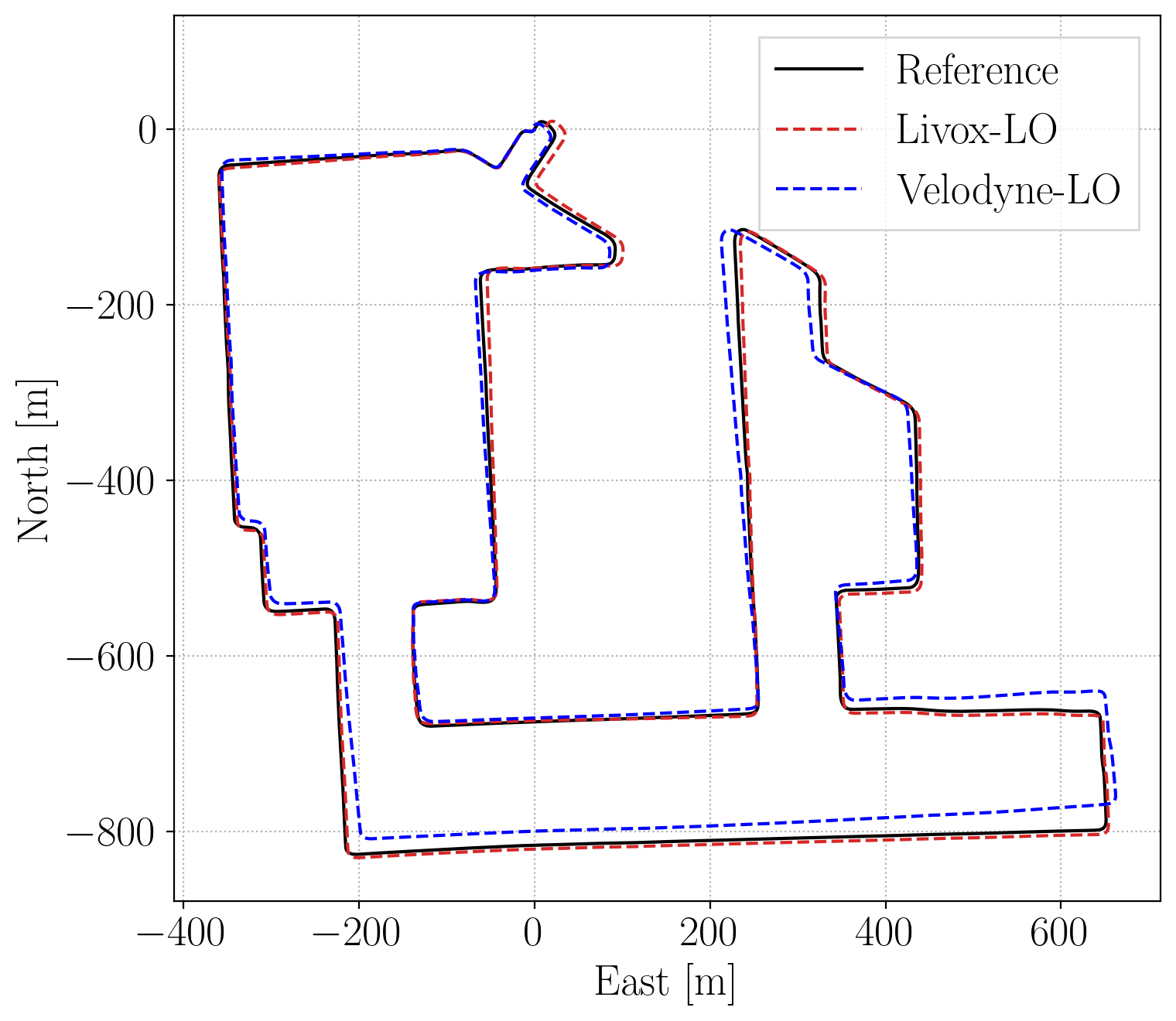}
        \caption{\texttt{Urban03}.}
        \label{fig:lo-urban03}
    \end{subfigure}
    \hspace{15pt}
    \begin{subfigure}{0.43\textwidth}
        \centering
        \includegraphics[height=0.8\linewidth]{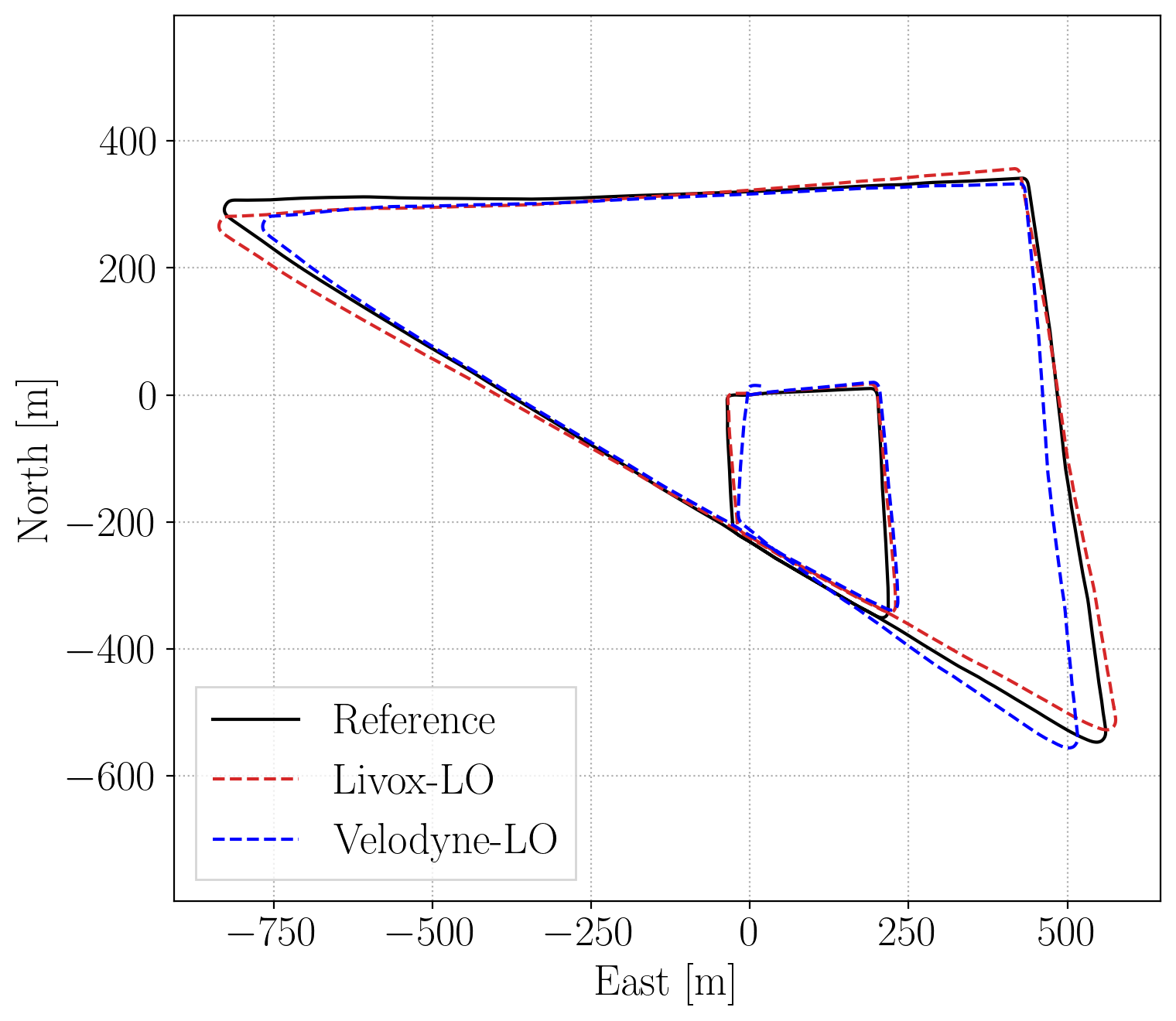}
        \caption{\texttt{Urban04}.}
        \label{fig:lo-urban04}
    \end{subfigure}
    \caption{\ac{lo} performance from the two \acp{lidar}.}
    \label{fig:lidar-benchmark}
\end{figure*}

\figurename~\ref{fig:lidar-benchmark} qualitatively demonstrates the performance of the estimated \ac{lo} solutions for the \texttt{Urban03} and \texttt{Urban04} trajectories. The solutions are plotted in 2D along the east and north axes, with the starting point set at coordinates $(0, 0)$. The precise and detailed spatial information obtained from the \ac{lidar} measurements results in trajectories that closely align with the reference. This demonstrates the ability the algorithm combined with \ac{lidar} measurement to capture the dynamics for the entirety of both urban trajectories.

Although the 2D plots in \figurename~\ref{fig:lidar-benchmark} suggest that both \acp{lidar} offer comparable \ac{lo} performance, with Livox-\ac{lo} showing slightly better alignment, \tablename~\ref{tab:lidar-performance} presents further insights. The \ac{ate} statistics for both position and rotation indicate a noticeable difference between the two systems. This is likely influenced by the vertical resolution of the Velodyne \ac{lidar} impacting the estimation of the vertical pose components. Regarding \ac{rpe}, Livox-\ac{lo} shows better translational accuracy, while the rotational performance varies between the two systems. Researchers are encouraged to further investigate the advantages of these cost-effective \ac{lidar} sensors, as well as the challenges posed by their varying specifications across different scenarios.

\begin{table}
\centering
\renewcommand{\arraystretch}{0.8}
\caption{Performance statistics from the two \acp{lidar} in both trajectories.}
\label{tab:lidar-performance}
\resizebox{\columnwidth}{!}{%
\begin{tabular}{@{}cccccc@{}}
\toprule
\multirow{2}{*}{\textbf{Sequence}} & \multirow{2}{*}{\textbf{Method}} & \multicolumn{2}{c}{\textbf{\ac{ate}}}               & \multicolumn{2}{c}{\textbf{\ac{rpe}}}               \\ \cmidrule(l){3-6} 
                                   &                                  & \textit{trans (m)} & \textit{rot (${}^\circ$)} & \textit{trans (m)} & \textit{rot (${}^\circ$)} \\ \midrule
\multirow{2}{*}{\texttt{Urban03 (D)}} & Livox-\ac{lo}    & 6.75   & 1.89  & 0.1114 & 0.2145 \\
                                  & Velodyne-\ac{lo} & 65.40  & 13.49 & 0.1425 & 0.1830 \\ \midrule
\multirow{2}{*}{\texttt{Urban04 (D)}} & Livox-\ac{lo}    & 23.61  & 2.79  & 0.1219 & 0.1136 \\
                                  & Velodyne-\ac{lo} & 109.16 & 22.41 & 0.2033 & 0.1343 \\ \bottomrule
\end{tabular}%
}
\end{table}

\subsection{Inertial Navigation System} \label{sec:inertial-navigation-system}
We employed the standard \ac{ins} algorithm to demonstrate the validity of the \ac{imu} data \cite{noureldin2012fundamentals}. The \ac{ins} algorithm processes linear acceleration and angular velocity measurements from each \ac{imu}, generating a distinct \ac{ins} solution. To mitigate the inherent drift in the \ac{ins}, the algorithm is complemented by the forward speed from the vehicle's odometer and the non-holonomic constraint for vehicular motion \cite{mounier2022online}. Additionally, initial biases in \ac{imu} measurements were estimated and corrected using data from a stationary period at the start of each trajectory, except the high-end \ac{imu}, where such corrections were not necessary.

\begin{figure*}
    \centering
    \begin{subfigure}{0.43\textwidth}
        \centering
        \includegraphics[height=0.8\linewidth]{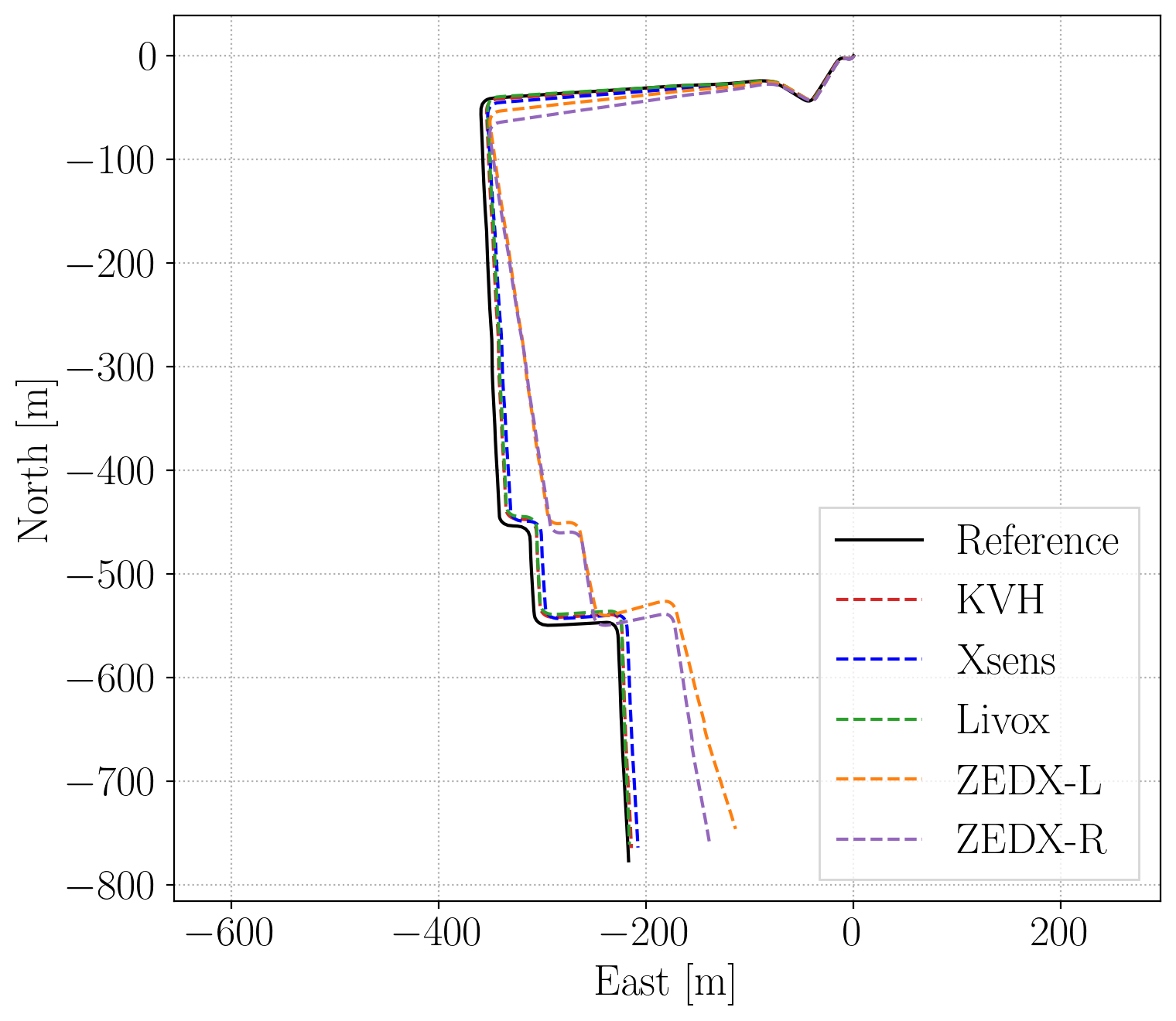}
        \caption{\texttt{Urban03}.}
        \label{fig:obms-urban03}
    \end{subfigure}
    \hspace{15pt}
    \begin{subfigure}{0.43\textwidth}
        \centering
        \includegraphics[height=0.8\linewidth]{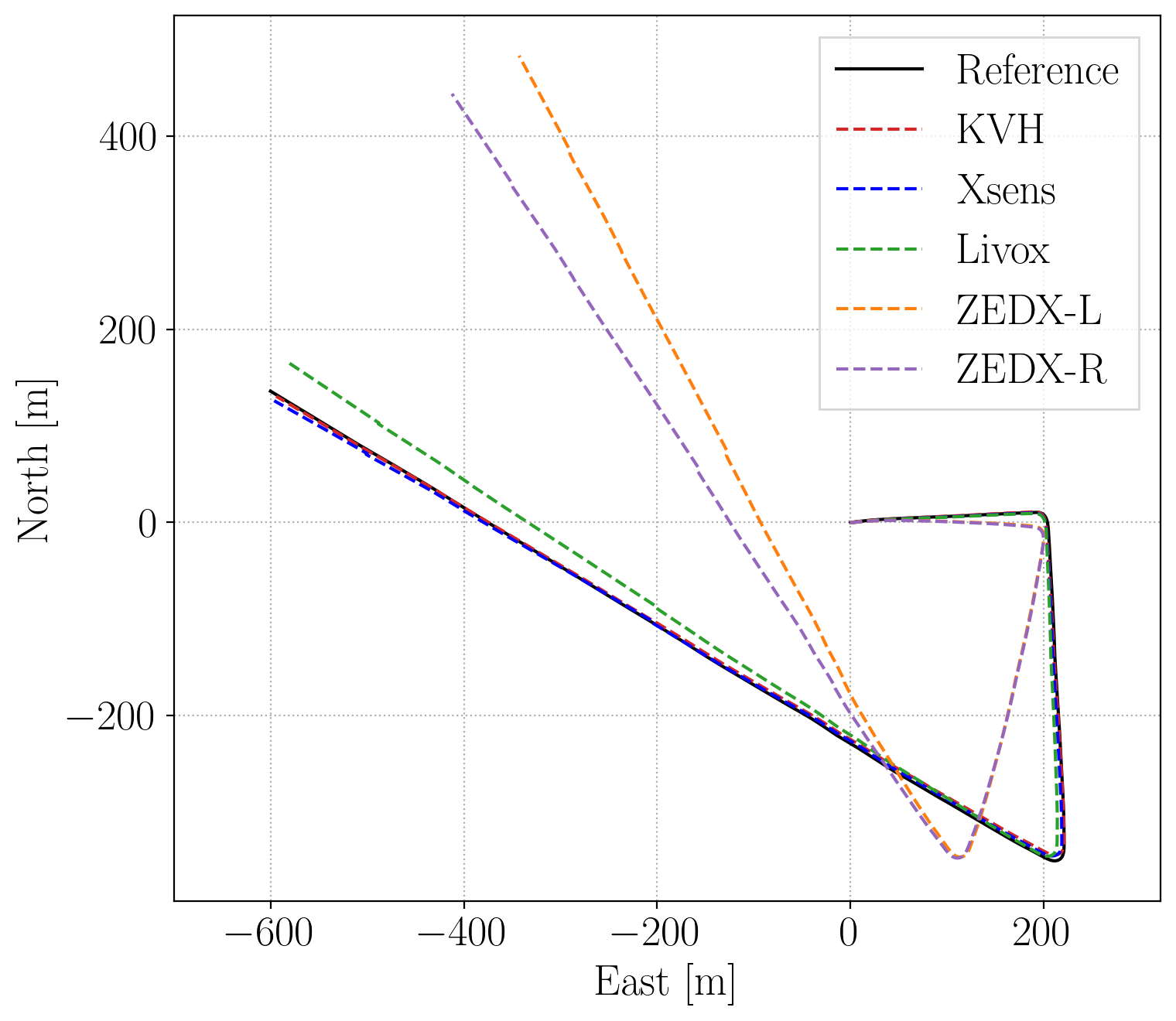}
        \caption{\texttt{Urban04}.}
        \label{fig:obms-urban04}
    \end{subfigure}
    \caption{\ac{ins} performance across different \acp{imu}.}
    \label{fig:imus-benchmark}
\end{figure*}

Given that some \acp{imu} exhibited significant drift, we limited the demonstration to 5-minute scenarios from the \texttt{Urban03} and \texttt{Urban04} trajectories. The performance of various \acp{imu} across these scenarios is presented in the 2D plots of \figurename~\ref{fig:imus-benchmark}. As anticipated, the KVH1750 \ac{imu} closely aligned with the reference trajectory in both scenarios, and the commercial \ac{imu} inside the Xsens unit also presented robust performance. The Livox built-in \ac{imu} demonstrated strong performance in both scenarios. In contrast, the \acp{imu} integrated within the ZED X cameras showed the least accurate performance, with considerable drift observed over extended periods.

\tablename~\ref{tab:imu-performance} provides a quantitative analysis of the \ac{imu} performance, with the KVH1750 \ac{imu} exhibiting superior \ac{ate} statistics in both translation and rotation, followed by the Livox and Xsens \acp{imu}. The \acp{imu} integrated into the cameras, however, clearly faced notable challenges resulting in significantly higher \ac{ate} values. The \ac{rpe} statistics reflect a similar trend, with the KVH \ac{imu} showing the best translation performance and the camera \acp{imu} recording the highest RPEs.

These results highlight the diverse range of \ac{imu} measurements our system offers, accommodating various levels of \ac{imu} quality from low-cost, built-in camera \acp{imu} to the high-end tactical-grade KVH1750 \ac{imu}. This diversity allows researchers and practitioners to explore and investigate the potential and limitations of different \ac{imu} grades.

\begin{table}
\centering
\renewcommand{\arraystretch}{0.8}
\caption{Performance statistics from all \acp{imu} in both trajectories.}
\label{tab:imu-performance}
\resizebox{\columnwidth}{!}{%
\begin{tabular}{@{}cccccc@{}}
\toprule
\multirow{2}{*}{\textbf{Trajectory}} & \multirow{2}{*}{\textbf{Method}} & \multicolumn{2}{c}{\textbf{\ac{ate}}}               & \multicolumn{2}{c}{\textbf{\ac{rpe}}}               \\ \cmidrule(l){3-6} 
                                   &                                  & \textit{trans (m)} & \textit{rot (${}^\circ$)} & \textit{trans (m)} & \textit{rot (${}^\circ$)} \\ \midrule
\multirow{5}{*}{\texttt{Urban03 (D)}} & KVH           & 7.01   & 0.65  & 0.0044 & 0.0347 \\
                                  & Xsens         & 11.61 & 1.11 & 0.0049 & 0.0348 \\
                                  & Livox         & 9.31  & 1.16 & 0.0045 & 0.0349 \\
                                  & ZEDX-L        & 43.21  & 10.58  & 0.0171 & 0.0363 \\
                                  & ZEDX-R        & 42.59  & 7.34  & 0.0121 & 0.0354 \\ \midrule
\multirow{5}{*}{\texttt{Urban04 (D)}} & KVH           & 10.69  & 0.96  & 0.0034 & 0.0132 \\
                                  & Xsens         & 13.72  & 1.74 & 0.0045 & 0.0073 \\
                                  & Livox         & 19.35  & 1.87 & 0.0047 & 0.0124 \\
                                  & ZEDX-L        & 205.35 & 26.18 & 0.0567 & 0.0082 \\
                                  & ZEDX-R        & 175.61 & 22.71 & 0.0494 & 0.0066 \\ \bottomrule
\end{tabular}%
}
\end{table}

\subsection{Monocular Visual Odometry} \label{sec:mvo}
\begin{figure*}
    \centering
    \begin{subfigure}{0.325\textwidth}
        \centering
        \includegraphics[width=\textwidth]{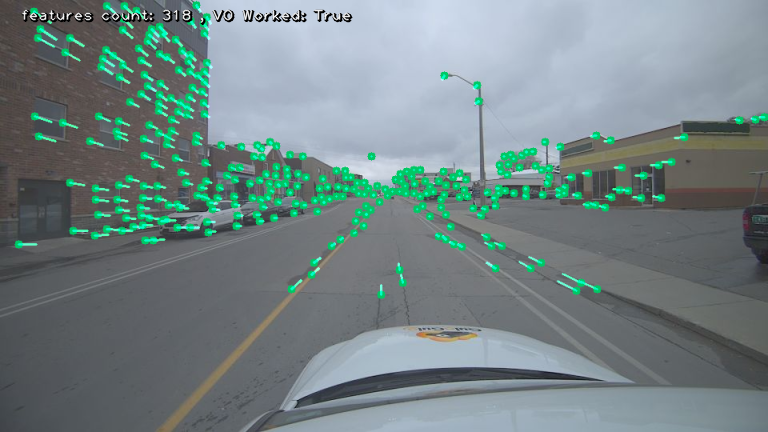}
        \caption{ZEDX-L.}
        \label{fig:zedxl-feat}
    \end{subfigure}
    \begin{subfigure}{0.325\textwidth}
        \centering
        \includegraphics[width=\textwidth]{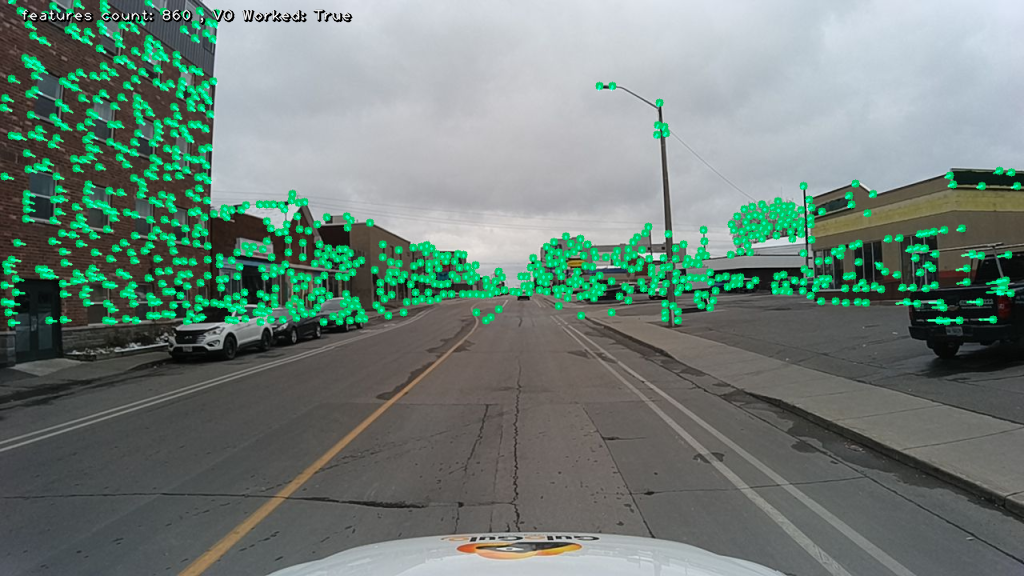}
        \caption{OAK.}
        \label{fig:oak-feat}
    \end{subfigure}
    \begin{subfigure}{0.325\textwidth}
        \centering
        \includegraphics[width=\textwidth]{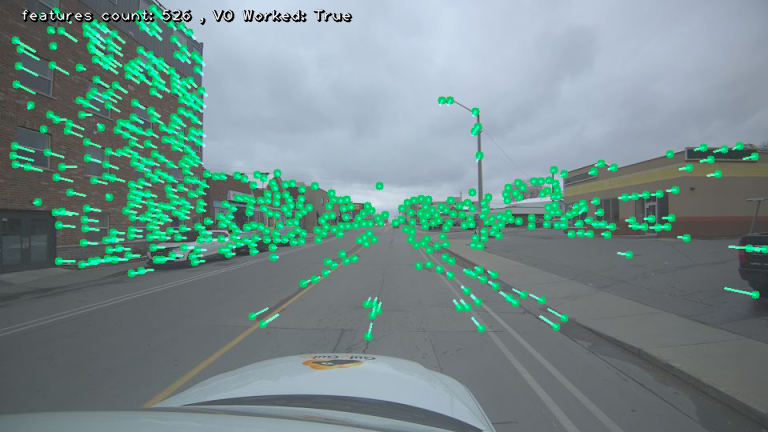}
        \caption{ZEDX-R.}
        \label{fig:zedxr-feat}
    \end{subfigure}
    \caption{Shi-Tomasi Features tracked across the three selected cameras.}  
    \label{fig:MVO-feats}
\end{figure*}

To validate some of the camera systems available in this dataset, we integrating cameras and inertial systems to perform \ac{mvo} on land vehicles~\cite{abdelaziz2020low,abdelaziz2024body}.
The method used relies on Shi-Tomasi features, provided by OpenCV's \textit{goodFeaturesToTrack} and pyramidal optical flow to track features, thus performing essential matrix estimation and recovering motion without having to perform the time-consuming feature matching operation. \figurename~\ref{fig:MVO-feats} shows the tracked features across the three cameras chosen for this demonstration.

\begin{figure*}
    \centering
    \begin{subfigure}{0.43\textwidth}
        \centering
        \includegraphics[height=0.8\linewidth]{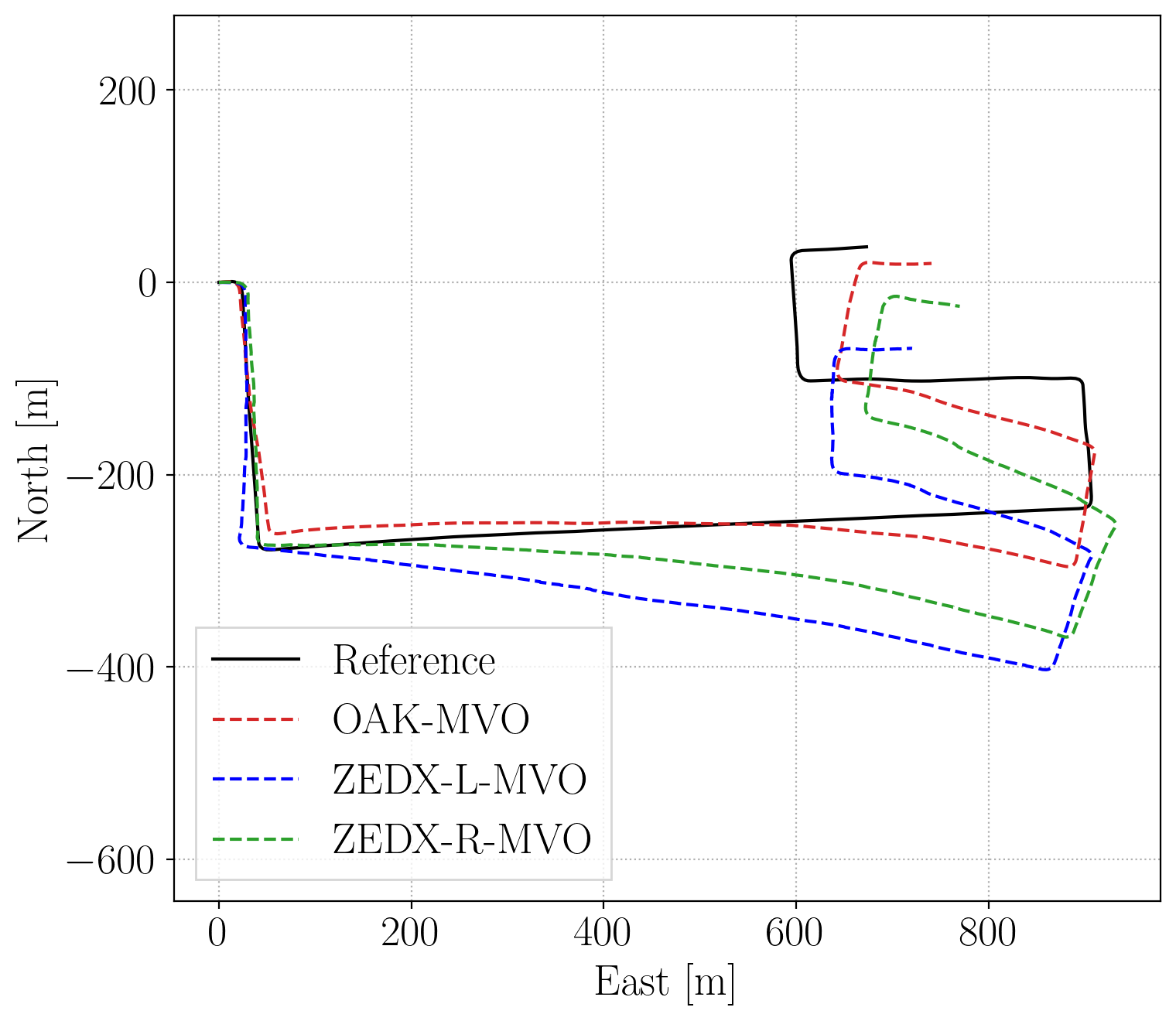}
        \caption{\texttt{Urban03}.}
        \label{fig:mvo-urban03}
    \end{subfigure}
    \hspace{15pt}
    \begin{subfigure}{0.43\textwidth}
        \centering
        \includegraphics[height=0.8\linewidth]{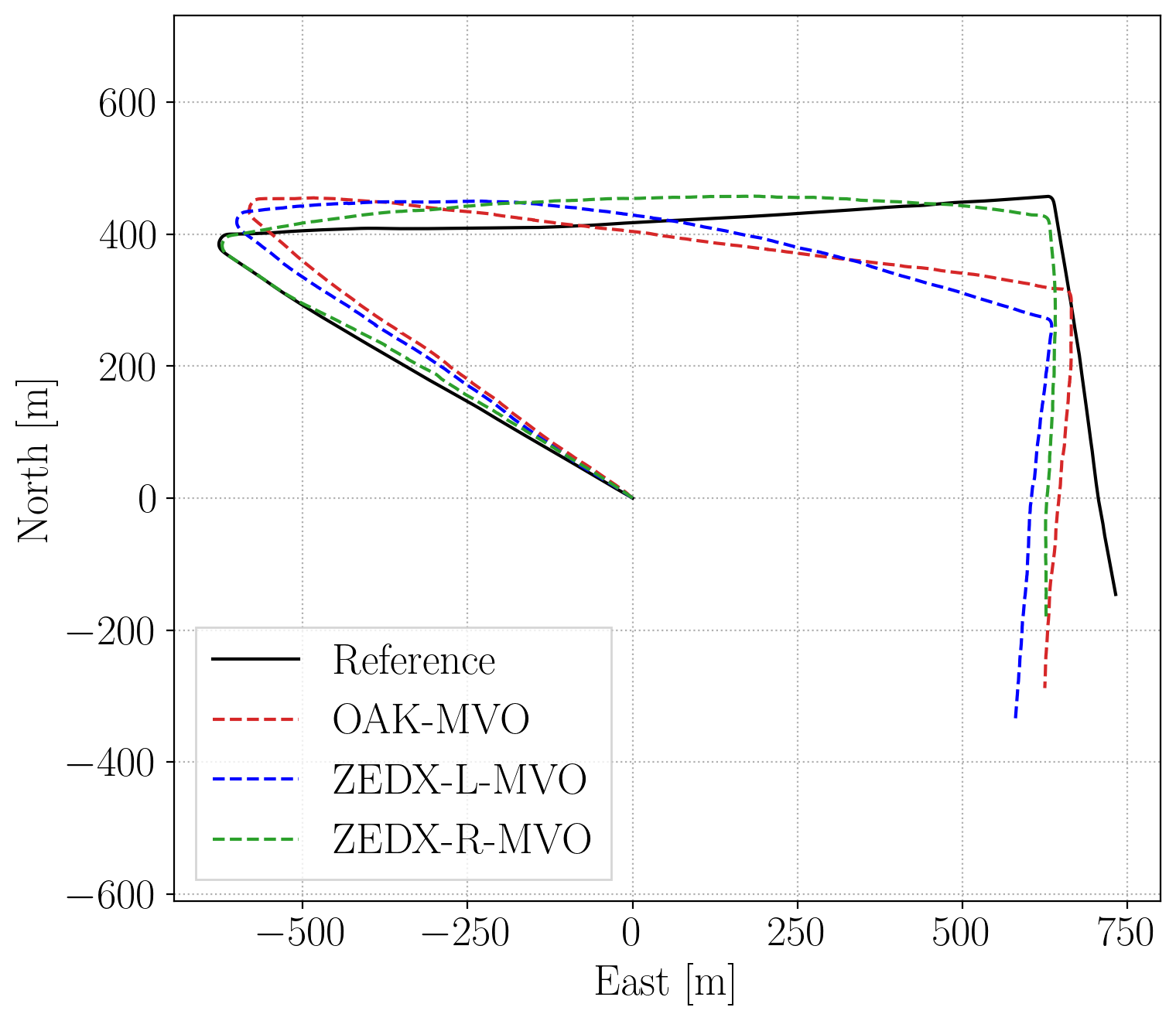}
        \caption{\texttt{Urban04}.}
        \label{fig:mvo-urban04}
    \end{subfigure}
    \caption{\ac{mvo} performance across different cameras.}
    \label{fig:vo-benchmark}
\end{figure*}

\figurename~\ref{fig:vo-benchmark} shows samples from the \ac{mvo} solutions, with translation scaled by the speed acquired from the odometer. Camera features are affected by moving objects in the scene; thus, the right camera consistently performs better than the left, where cars commonly pass in the opposing direction.  

\tablename~\ref{tab:camera-performance} shows the \ac{mvo} performance across the selected cameras. Consistent with evidence in the computer vision literature, standalone cameras cannot provide long-term positioning solutions. However, we believe that by providing data from a multi-view platform, our dataset will contribute to recent multi-modal data fusion trends in computer vision.

\begin{table}
\centering
\renewcommand{\arraystretch}{0.8}
\caption{Performance statistics from different cameras in both trajectories.}
\label{tab:camera-performance}
\resizebox{\columnwidth}{!}{%
\begin{tabular}{@{}cccccc@{}}
\toprule
\multirow{2}{*}{\textbf{Trajectory}} & \multirow{2}{*}{\textbf{Method}} & \multicolumn{2}{c}{\textbf{\ac{ate}}}               & \multicolumn{2}{c}{\textbf{\ac{rpe}}}               \\ \cmidrule(l){3-6} 
                                   &                                  & \textit{trans (m)} & \textit{rot (${}^\circ$)} & \textit{trans (m)} & \textit{rot (${}^\circ$)} \\ \midrule
\multirow{2}{*}{\texttt{Urban03 (D)}} & OAK-\ac{mvo}    & 39.74 & 10.63  &  0.0726 &  0.1087 \\
                                  & ZEDX-L-\ac{mvo} & 108.63 &  9.34  &  0.1201 &  0.1341 \\
                                  & ZEDX-R-\ac{mvo} & 82.00 &  11.84  &  0.1259 &  0.1295 \\ \midrule
\multirow{2}{*}{\texttt{Urban04 (D)}} & OAK-MVO    & 91.83 &  4.74  &  0.1309 &  0.0875 \\
                                  & ZEDX-L-\ac{mvo} & 113.06 &  9.14  &  0.1921 &  0.1343 \\
                                  & ZEDX-R-\ac{mvo} & 35.37 &  5.60  &  0.1651 &  0.1420 \\ \bottomrule
\end{tabular}%
}
\end{table}

\subsection{Stereo Visual Simultaneous Localization and Mapping} \label{sec:svSLAM}
We also demonstrate the applicability of the cameras for stereo \ac{vslam}. In this application, we are interested in localizing the camera while building a map of the environment~\cite{mur2017orb}. The process begins with a map initialization, where a 3D map is generated from a pair of stereo images, with the left image stored as the first keyframe and 3D points derived from the disparity map.

In the subsequent stage, i.e., tracking, the camera's pose is estimated by matching features in the left image to the last keyframe and refined by tracking the disparity, ensuring accuracy in the estimated trajectory. Next, new map points are added if the current left image is identified as a keyframe, with bundle adjustment minimizing reprojection errors by refining both the camera's pose and 3D points. The final stage ensures long-term map consistency by detecting and optimizing loop closures, refining the camera's pose and 3D map to correct any drift.

\begin{figure}
    \centering
    \includegraphics[width=\linewidth]{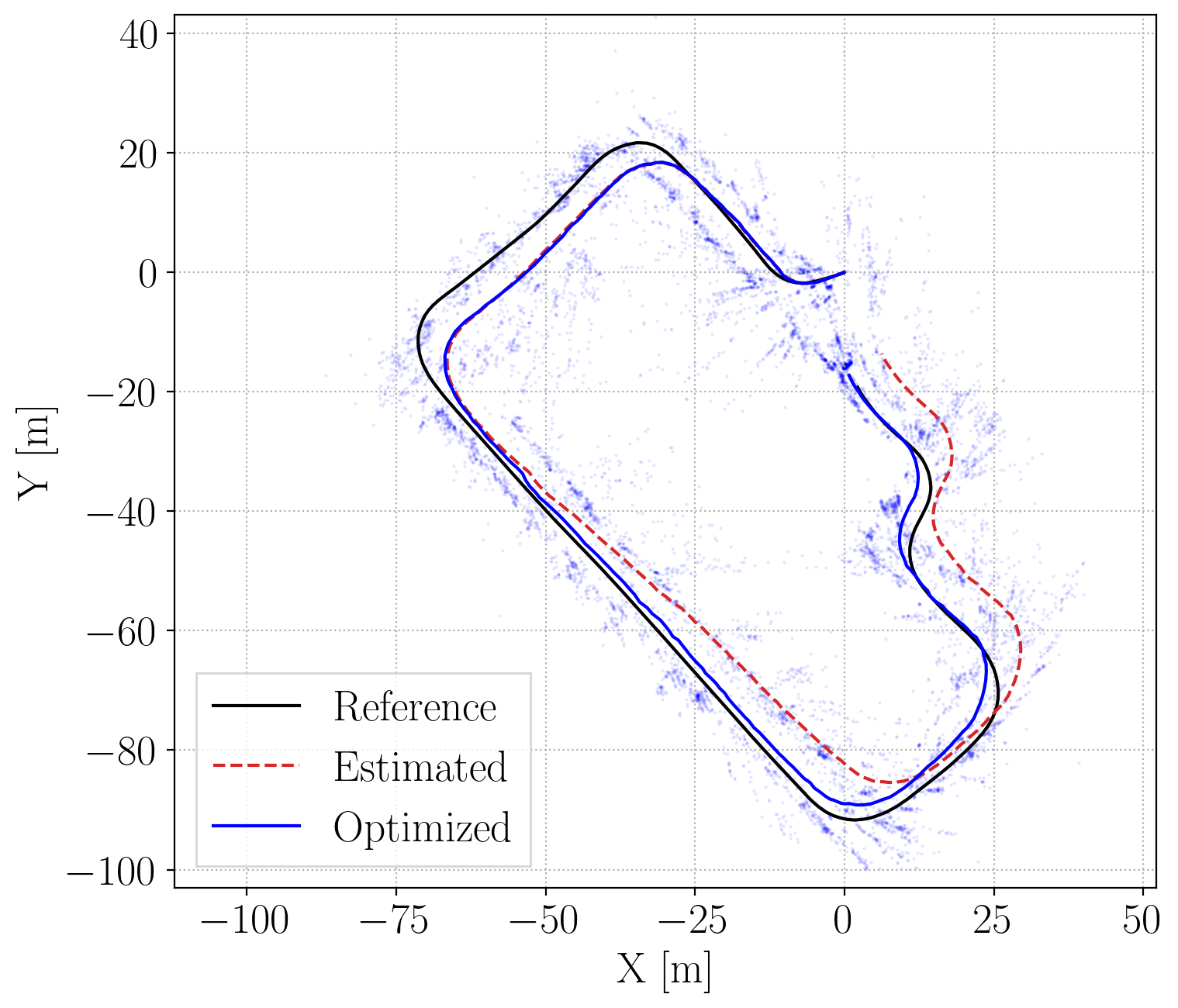}
    \caption{Performance of Stereo \ac{vslam} in \texttt{Indoor02}.}
    \label{fig:sv-slam}
\end{figure}

\figurename~\ref{fig:sv-slam} presents the results obtained with the ZED X (left) stereo camera. It can be seen that the method captures the vehicle's dynamics, albeit with some scale error. The stereo camera has a wide field of view and a short baseline, which can affect the accuracy of depth estimation and, consequently, the scale of the estimated poses. The blue dots represent the detections included in the map. This scenario, along with all indoor trajectories, offers many research opportunities, especially since the ground truth map of the garage is provided.

\tablename~\ref{tab:vslam-performance} presents the evaluation metrics for both the estimated and optimized trajectories. The optimization process noticeably improved the \ac{ate} in translation, reducing it from 6.56 meters to 3.83 meters. However, this optimization negatively impacted the rotation metrics, which could be attributed to the lack of loops in the tested scenario. Overall, the results highlight the potential of the cameras in the platform for application to various advanced methods. Users can create stereo rigs using the available cameras and take advantage of longer baselines.

\begin{table}
\centering
\renewcommand{\arraystretch}{0.8}
\caption{\ac{vslam} performance statistics in indoor settings using the ZED X (Left) stereo camera.}
\label{tab:vslam-performance}
\resizebox{\columnwidth}{!}{%
\begin{tabular}{@{}cccccc@{}}
\toprule
\multirow{2}{*}{\textbf{Trajectory}} & \multirow{2}{*}{\textbf{Method}} & \multicolumn{2}{c}{\textbf{\ac{ate}}}               & \multicolumn{2}{c}{\textbf{\ac{rpe}}}               \\ \cmidrule(l){3-6} 
                                     &                                  & \textit{trans (m)} & \textit{rot (${}^\circ$)} & \textit{trans (m)} & \textit{rot (${}^\circ$)} \\ \midrule
\multirow{2}{*}{\texttt{Indoor02}} & Estimated    & 6.56 & 2.31 & 0.1534 & 0.3103 \\
                                   & Optimized     & 3.83 & 2.82 & 0.1586 & 0.6897 \\
                                  \bottomrule
\end{tabular}%
}
\end{table}

\subsection{RADAR Odometry and Map Registration}
The validity of the \ac{radar} data from the four Smartmicro \acp{radar} is demonstrated through the implementation of two \ac{radar}-based localization methods applied to an indoor trajectory: \ac{rio} and a \ac{rmr} positioning solution. The \ac{rio} method leverages the Doppler velocities and azimuths of targets returned by the \ac{esr} to estimate the instantaneous forward velocity of the vehicle following the procedure outlined in \cite{kellner_instantaneous_2013}, which is then used as forward speed updates in an \ac{ins} mechanization as in \cite{dawson2023integrated}. The \ac{rmr} method employs the \ac{icp} algorithm to register \ac{radar} scans to a map of the environment for positioning corrections, following methods presented in \cite{dawson2023integrated}.

\begin{figure}
    \centering
    \includegraphics[width=\linewidth]{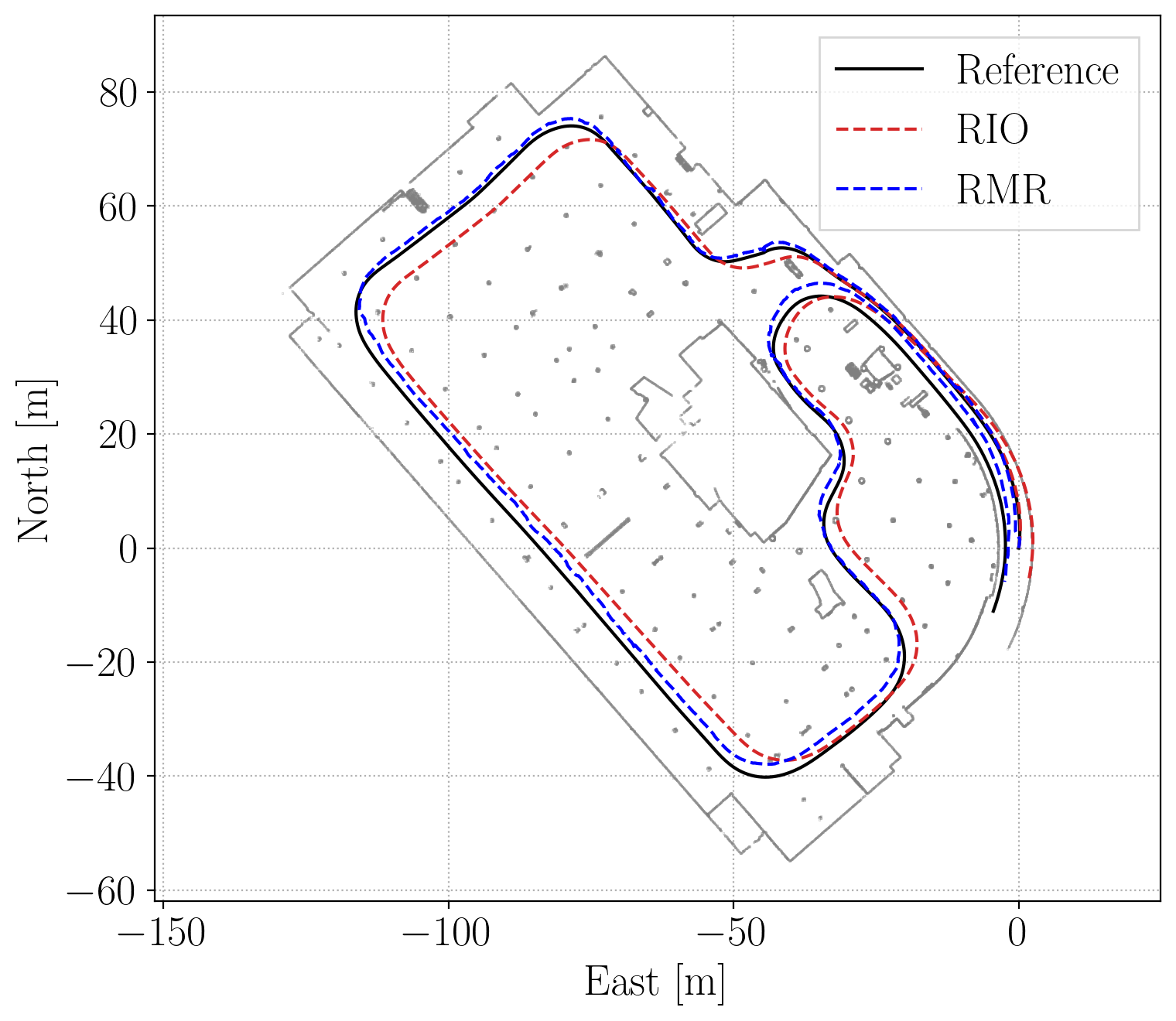}
    \caption{\ac{rio} vs. \ac{rmr} performance in \texttt{Indoor02} Trajectory.}
    \label{fig:radar-indoor02}
\end{figure}

\figurename~\ref{fig:radar-indoor02} shows the performance of both the \ac{rio} and the \ac{rmr} solutions in an indoor parking garage, compared to the ground truth solution. A quantitative analysis of the solutions is provided in \tablename~\ref{tab:radar-performance}. \ac{rio}, as a dead reckoning solution, drifts throughout the course of the 2.7-minute trajectory, while \ac{rmr} maintains bounded positioning errors throughout the entire experiment.

Localization using automotive \acp{radar} remains a challenging task due to their comparatively low resolution and high noise and clutter levels. However, these results demonstrate the promise of automotive \ac{radar} as an aiding sensor in automotive localization. The dataset facilitates research into the emerging possibilities and challenges relating to automotive \ac{radar} as a localization sensor. 

\begin{table}
\renewcommand{\arraystretch}{0.8}
\centering
\caption{\ac{radar} performance statistics in indoor settings.}
\label{tab:radar-performance}
\resizebox{\columnwidth}{!}{%
\begin{tabular}{@{}cccccc@{}}
\toprule
\multirow{2}{*}{\textbf{Trajectory}} & \multirow{2}{*}{\textbf{Method}} & \multicolumn{2}{c}{\textbf{\ac{ate}}}               & \multicolumn{2}{c}{\textbf{\ac{rpe}}}               \\ \cmidrule(l){3-6} 
                                   &                                  & \textit{trans (m)} & \textit{rot (${}^\circ$)} & \textit{trans (m)} & \textit{rot (${}^\circ$)} \\ \midrule
\multirow{2}{*}{\texttt{Indoor02}} & \ac{rio}    & 3.79 & 2.51 & 0.0170 & 0.0263 \\
                                   & \ac{rmr}    & 1.89 & 3.37 & 0.1065 & 0.3650 \\
                                  \bottomrule
\end{tabular}%
}
\end{table}

\subsection{Forward Speed using RADAR} \label{sec:forward-speed-radar}
The \ac{radar} mounted on the vehicle's front bumper can be used to estimate the vehicle's forward speed. As the sensor faces the ground at an angle, it measures the relative speed between the vehicle and the ground. Since the ground is always available and stationary, it is possible to infer the vehicle's forward speed by knowing the sensor's mounting angle.

The challenge lies in the accurate computation of the mounting angle, as vehicles have different shapes and mounting positions. To solve this problem, some researchers proposed a deep-learning-based method to calibrate the mounting angle and denoise the signal~\cite{de2023novel}. The system is based on a denoising autoencoder and is trained using a reference system.

\begin{figure}
    \centering
    \includegraphics[width=\linewidth]{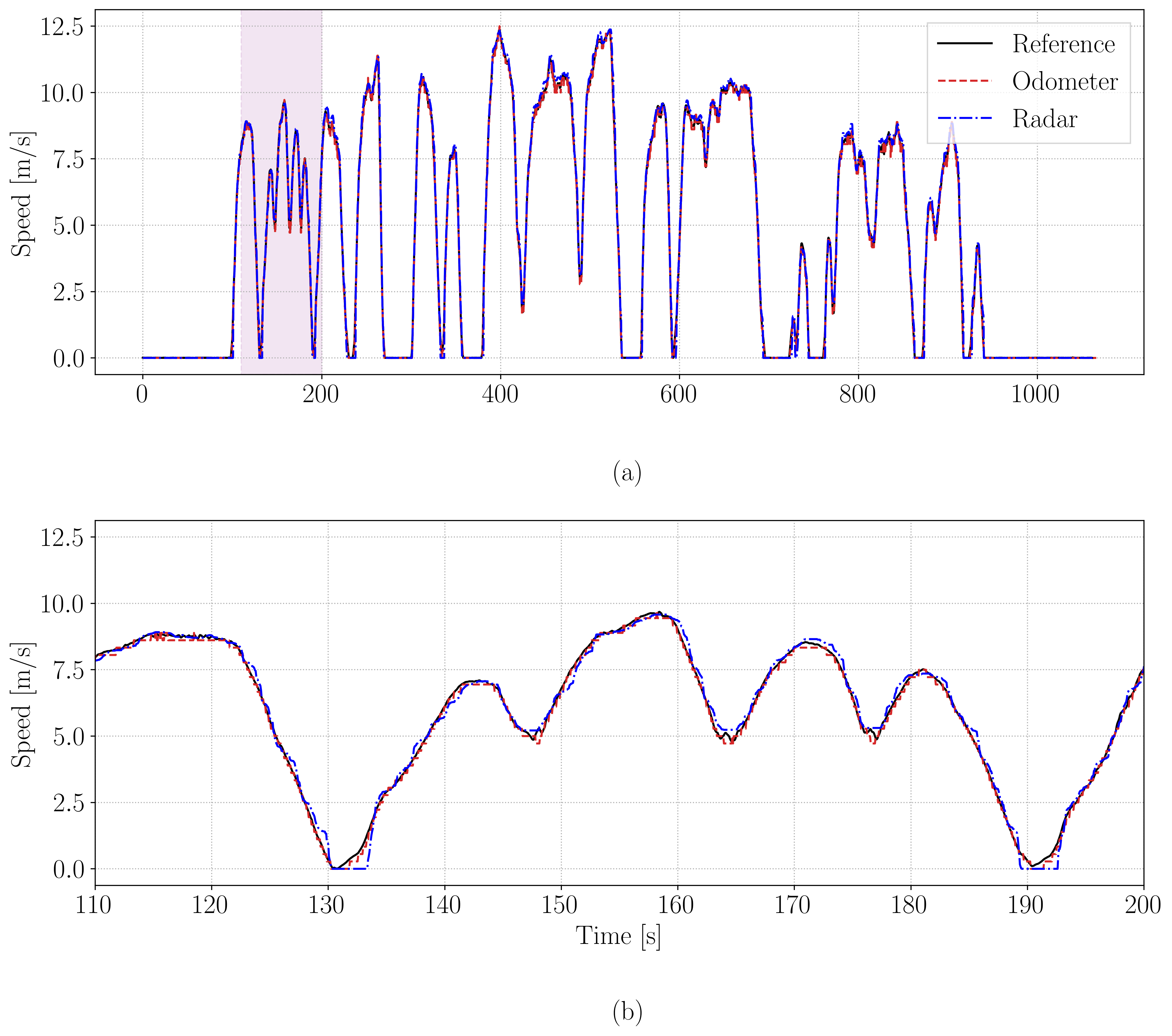}
    \caption{Vehicle's forward speed comparison.}
    \label{fig:forward-speed-radar}
\end{figure}

\figurename~\ref{fig:forward-speed-radar} demonstrates such an approach. The blue dashed line represents the results of the denoising autoencoder, which was trained using the reference system, i.e., Novatel PwrPak7-E1 + KVH1750 \ac{imu}. We compare the \ac{radar}-based estimation to the reference system and the odometer. As seen in the highlighted area magnified in \figurename~\ref{fig:forward-speed-radar}(b), the model performs well and captures most of the dynamics.

The \ac{radar}-based estimation obtained a root-mean-squared error of 0.24 m/s when compared to the reference system, while the odometer-based estimation achieved 0.11 m/s. Practitioners can implement different methods or strategies to improve the results and generate a reliable and redundant source of forward speed measurement.

\section{Conclusion} \label{sec:conclusion}

This paper introduces the \ac{navinst} dataset, a comprehensive collection of land vehicle trajectories recorded under various lighting conditions and across diverse urban and indoor environments. Designed to support the research community, our open-source dataset is fully integrated with \ac{ros}, offering both accessibility and ease of use. The primary objective of this dataset is to contribute to the advancement of navigation and positioning research, particularly in the development and validation of robust algorithms for future commercial autonomous vehicles. However, we also encourage researchers to explore and adapt this dataset for a wide range of other research applications beyond autonomous navigation.

The dataset provides a plethora of sensors, including multiple-grade \acp{imu}, \ac{gnss}, monocular and stereo cameras, mechanical and solid-state \acp{lidar}, automotive \acp{radar}, and onboard vehicle measurements. We demonstrate the validity and effectiveness of our sensor data through comprehensive demonstrations employing various navigation algorithms. These demonstrations underscore the potential of the dataset to facilitate a vast range of navigation-related research endeavours. This project is an ongoing effort, and future work will focus on further enriching the \ac{navinst} dataset. We plan to investigate the inclusion of additional sensors, explore new environments, and expand the dataset's applicability to a broader range of scenarios. These enhancements will be guided by feedback from the research community, ensuring that the dataset continues to evolve to better support navigation and positioning research.

\section*{Acknowledgment}
This research is supported by grants from the Natural Sciences and Engineering Research Council of Canada (NSERC) under grant numbers RGPIN-2020-03900 and ALLRP-560898-20. The authors also thank Tristan Redish and Marc Adam for their invaluable support during the execution of this project and for coordinating work between different technical departments. In addition, we extend our gratitude to past members of the NavINST Lab, whose foundational experiments and insightful discussions have significantly contributed to the evolution of this work. Finally, we thank Micro Engineering Tech Inc. (METI) for their technical support and for preparing and providing the 3D maps of the indoor garages.

\ifCLASSOPTIONcaptionsoff
  \newpage
\fi

\end{document}

%% file: acronyms.tex
\usepackage{acro}
\DeclareAcronym{navinst}{
  short = NavINST,
  long  = Navigation and Instrumentation}
\DeclareAcronym{gnss}{
  short = GNSS,
  long  = global navigation satellite system}
  \DeclareAcronym{lidar}{
  short = LiDAR,
  long  = light detection and ranging}
  \DeclareAcronym{radar}{
  short = RADAR,
  long  = radio detection and ranging}
\DeclareAcronym{av}{
  short = AV,
  long  = autonomous vehicle}
\DeclareAcronym{imu}{
  short = IMU,
  short-plural-form = IMUs,
  long  = inertial measurement unit,
  long-plural-form = inertial measurement units}
\DeclareAcronym{mems}{
  short = MEMS,
  long  = micro-electro-mechanical systems}
\DeclareAcronym{ins}{
  short = INS,
  long  = inertial navigation system} 
\DeclareAcronym{kf}{
  short = KF,
  long  = Kalman filter}
\DeclareAcronym{lo}{
  short = LO,
  long  = LiDAR odometry}
\DeclareAcronym{lio}{
  short = LIO,
  long  = LiDAR-inertial odometry}
\DeclareAcronym{vins}{
  short = VINS,
  long  = visual-inertial navigation system}
\DeclareAcronym{enu}{
  short = ENU,
  long  = east-north-up}
\DeclareAcronym{ros}{
  short = ROS,
  long  = robot operating system}
\DeclareAcronym{rtk}{
  short = RTK,
  long  = real-time kinematics}
\DeclareAcronym{esr}{
  short = ESR,
  short-plural-form = ESRs,
  long  = electronically scanning RADAR,
  long-plural-form = electronically scanning RADARs}
\DeclareAcronym{rio}{
  short = RIO,
  long  = RADAR-Inertial Odometry}
\DeclareAcronym{rmr}{
  short = RMR,
  long = RADAR-to-map registration
}
\DeclareAcronym{slam}{
  short = SLAM,
  long = simultaneous localization and mapping 
}
\DeclareAcronym{lmr}{
  short = LMR,
  long = LiDAR-to-map registration
}
\DeclareAcronym{mvo}{
  short = MVO,
  long = monocular visual odometry
}
\DeclareAcronym{usb}{
  short = USB,
  long = universal serial bus
}
\DeclareAcronym{gmsl}{
  short = GMSL,
  long = gigabit multimedia serial link
}
\DeclareAcronym{ntp}{
  short = NTP,
  long = network time protocol
}
\DeclareAcronym{ppk}{
  short = PPK,
  long = post-processed kinematic
}
\DeclareAcronym{ate}{
  short = ATE,
  long = absolute trajectory error
}
\DeclareAcronym{rpe}{
  short = RPE,
  long = relative pose error
}
\DeclareAcronym{icp}{
  short = ICP,
  long = iterative closest point
}
\DeclareAcronym{vslam}{
  short = VSLAM,
  long = visual simultaneous localization and mapping
}
